\documentclass[10pt,twocolumn,letterpaper]{article}

\usepackage[pagenumbers]{iccv} % To force page numbers, e.g. for an arXiv version
\DeclareRobustCommand{\name}{xxx}

\definecolor{iccvblue}{rgb}{0.21,0.49,0.74}
\usepackage[pagebackref,breaklinks,colorlinks,allcolors=iccvblue]{hyperref}
\usepackage{tabularx}
\usepackage{graphicx}
\usepackage{booktabs}
\usepackage{graphicx}

\usepackage{subcaption}
\usepackage{amsmath}
\usepackage{amssymb}
\usepackage{booktabs}
\usepackage{multirow}
\usepackage{tabularx}
\usepackage{enumitem}
\usepackage{gensymb}
\usepackage{colortbl}
\usepackage{caption}
\usepackage{makecell}
\usepackage{float} % For the H option
\usepackage{placeins}
\usepackage{wrapfig}
\usepackage{amssymb}
\usepackage{caption} 
\usepackage{tcolorbox}
\usepackage{adjustbox} % 允许自动缩放
\usepackage{lipsum} % 示例文本，可删除
\usepackage{pgffor} % 方便循环生成子图
\usepackage{pifont}% http://ctan.org/pkg/pifont
\definecolor{mygray}{gray}{.95}
\definecolor{mycyan}{cmyk}{.1,0,0,0}
\definecolor{darkgreen}{rgb}{0.0, 0.5, 0.0}

\usepackage[dvipsnames]{xcolor}
\definecolor{cjcolor}{RGB}{255,0,0}

\DeclareRobustCommand{\name}{RoboTron-Sim }

\newcommand{\cmarkg}{\ding{51}}%
\newcommand{\xmarkg}{\textcolor{lightgray}{\ding{55}}}%
\def\paperID{9411} % *** Enter the Paper ID here
\def\confName{ICCV}
\def\confYear{2025}

\title{\textit{RoboTron-Sim}: Improving Real-World Driving via Simulated Hard-Case}

\author{
  Baihui Xiao$^{1\dagger}$ \quad Chengjian Feng$^{1\dagger}$ \quad Zhijian Huang$^{1,2}$\\
  Feng Yan$^{1}$ \quad Yujie Zhong$^{1}$ \quad Lin Ma$^{1\ddagger}$
  \\
  \\
  $^{1}$Meituan ~ $^{2}$Shenzhen Campus of Sun Yat-sen University \\
  \url{https://stars79689.github.io/RoboTron-Sim}
}
\newcommand\blfootnote[1]{%
\begingroup
\renewcommand\thefootnote{}\footnote{#1}%
\addtocounter{footnote}{-1}%
\endgroup
}

\usepackage{xparse}
\usepackage{amsmath, amssymb, mathrsfs}
\begin{document}
\maketitle

\blfootnote{$\dagger$ Equal contribution. $\ddagger$ Corresponding author.}

\begin{abstract}
Collecting real-world data for rare high-risk scenarios, long-tailed driving events, and complex interactions remains challenging, leading to poor performance of existing autonomous driving systems in these critical situations.
In this paper, we propose \textbf{\name{}}that improves real-world driving in critical situations by utilizing simulated hard cases.
First, we develop a simulated dataset called Hard-case Augmented Synthetic Scenarios (HASS), which covers 13 high-risk edge-case categories, as well as balanced environmental conditions such as day/night and sunny/rainy.
Second, we introduce Scenario-aware Prompt Engineering (SPE) and an Image-to-Ego Encoder (I2E Encoder) to enable multimodal large language models to effectively learn real-world challenging driving skills from HASS, via adapting to environmental deviations and hardware differences between real-world and simulated scenarios.
Extensive experiments on nuScenes show that \name{}improves driving performance in challenging scenarios by $\sim$\textbf{50\%}, achieving state-of-the-art results in real-world open-loop planning.
Qualitative results further demonstrate the effectiveness of \name{}in better managing rare high-risk driving scenarios.

\end{abstract}    
\section{Introduction}
\label{sec:intro}

% 数据（hard case、long-tailed、risk）采集困难（场景和数量），在某些场景里表现差，仿真数据能够解决这个问题。我们尝试传统方法直接混合真实和仿真数据效果并不好，而多模态大模型展示出强大的泛化能力，为解决这个问题打开了新的机会。本文合成/采样一个仿真数据集、搭建了一个多模态planning大模型、prompt和tokenizer改进

% 端到端自动驾驶能够以数据驱动

The landscape of autonomous driving (AD) has seen great breakthroughs in recent years, driven primarily by the development of end-to-end AD systems~\cite{chen2024survey,yang2023survey,zhang2024survey, wen2023dilu, xu2024drivegpt4,huang2025making,cui2024drive, shao2024lmdrive,hu2023planning}.
These systems represent a paradigm shift from traditional module-based approaches, which separate perception~\cite{huang2021bevdet,liang2022bevfusion,liu2023bevfusion}, prediction~\cite{gu2023vip3d,gao2020vectornet,da2022path}, and planning~\cite{scheel2022urban,sadat2020perceive,gao2022cola} into distinct components. 
Instead, end-to-end AD synthesizes these functionalities into a unified, integrated framework, utilizing the robustness of foundational models~\cite{achiam2023gpt,yang2023llm4drive,brown2020language} and extensive datasets~\cite{ding2024holistic,li2024automated, marcu2024lingoqa} to catalyze innovation.
This data-driven approach not only enhances scalability~\cite{huang2025making,jia2024bench2drive} but also promises continuous improvement as incremental data integration~\cite{wu2023policy,yang2023llm4drive}. 
Thus, end-to-end AD is positioned to redefine the future of autonomous mobility by offering a more adaptive solution to complex driving scenarios.

% In the era of data-driven innovation, the availability of high-quality datasets plays a pivotal role. 
% However, in specific scenarios such as hard cases, long-tailed distributions, and risk-related contexts, data collection often encounters substantial challenges.

% \begin{figure}[t]
%     \centering
%     \begin{subfigure}[b]{0.48\textwidth}
%         \includegraphics[width=\textwidth]{env11.png}
%         \subcaption{Performance comparison of VAD across different datasets.}
%         \label{fig:vad_performance}
%     \end{subfigure}
%     \hfill
%     \begin{subfigure}[b]{0.48\textwidth}
%         \includegraphics[width=\textwidth]{env10.png} 
%         \subcaption{Comparison between baseline and \name{}.}
%         \label{fig:simboost_performance}
%     \end{subfigure}
    
%     \label{fig:intro_case}
% \end{figure}

\begin{figure}
    \centering
    \includegraphics[width=1\linewidth]{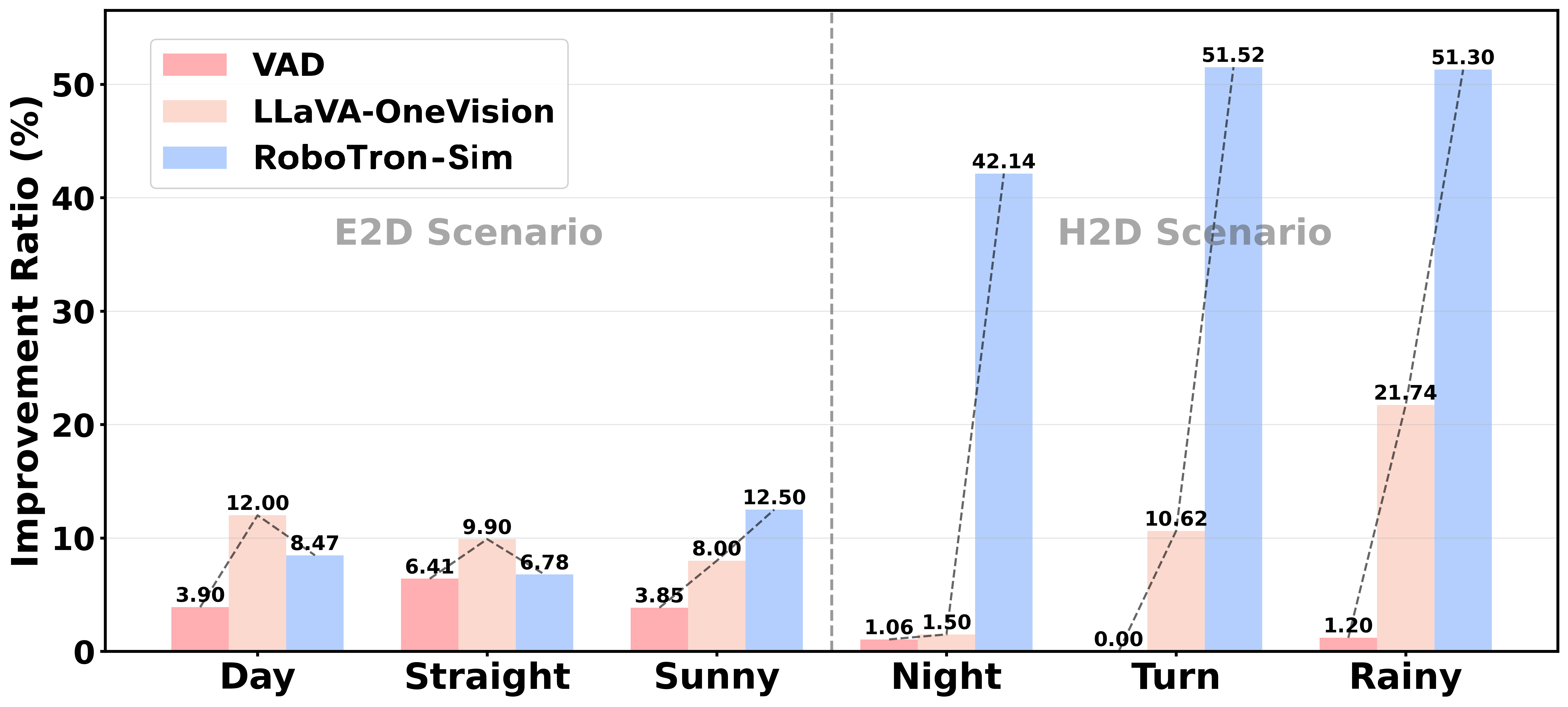}
    \vspace{-6mm}
    \caption{\name{}achieves stronger improvements in real-world driving capabilities by leveraging simulated hard-case data.
    We perform representative methods to evaluate the improvements from simulated data. Results show traditional method VAD achieves minor gains while LLaVA-OneVision struggles to improve performance in challenging scenarios. \name{}achieves over 50\% improvement in hard-to-drive scenarios.
    }
    \vspace{-6mm}
    \label{fig:intro_img}
\end{figure}

In the era of data-driven innovation, the availability of high-quality datasets~\cite{caesar2020nuscenes,sun2020waymo} plays a pivotal role in technological advancement. 
Current mainstream AD systems like UniAD~\cite{hu2023planning} and VAD~\cite{jiang2023vad}, which employ end-to-end architectures, remain constrained by supervised learning paradigms requiring fine-grained annotations such as 3D bounding boxes for agents and semantic segmentation of lane markings. 
% This stands in stark contrast to the development trajectory of large language models (LLMs) that leverage vast amounts of nearly free web-scale text data. 
The exorbitant cost of annotation has become a critical bottleneck in scaling these end-to-end methods, while the inherent long-tail distribution~\cite{jain2021autonomy} challenge in AD tasks exacerbates the data dilemma – most collected data consists of trivial scenarios like straight-road driving~\cite{li2024ego}, with safety-critical cases being extremely scarce. 
% This data imbalance significantly limits the practical effectiveness of data-driven approaches, particularly in long-tail distributions involving complex edge cases and risk-sensitive scenarios where data collection faces substantial challenges. 
This distribution shift significantly limits the practical effectiveness of data-driven approaches, particularly in complex edge cases and risk-sensitive scenarios where data collection faces substantial challenges.
Given the scarcity of real-world data in these critical areas, Methods~\cite{jia2024bench2drive,shao2024lmdrive} have increasingly turned to synthetic data as a potential solution. 
Simulation environment~\cite{dosovitskiy2017carla} offers the ability to generate diverse and controlled datasets, including rare and high-risk scenarios that are difficult to capture in the real world. 
These constraints raise a fundamental research question: \textit{Can simulation-generated datasets be effectively leveraged to enhance model performance in real-world scenarios?}

% While the cost-effectiveness of data acquisition in simulation environments enables promising paradigms for training with synthetic data prior to real-world deployment, conventional approaches that simply combine real-world samples with simulator-generated challenging scenarios yield limited performance improvements despite increased data volume.
While simulation data is cost-effective and enables training with synthetic challenging scenarios, conventional approaches (\eg~VAD) that simply mix real-world and simulated data achieve limited performance gains, even with larger datasets.
This constraint primarily stems from the persistent simulation-to-reality (Sim2Real) domain gap, as evidenced by our experimental observations in Fig.~\ref{fig:intro_img}, where models demonstrate inadequate capability in transferring knowledge acquired from synthetic scenarios to authentic driving contexts. 
% The fundamental challenge lies in the inherent discrepancies between simulated sensory inputs (e.g., simplified physical interactions) and their real-world counterparts, which hinder effective cross-domain knowledge transfer. Current methods that naively augment training datasets with simulated samples without addressing these domain-specific characteristics fail to establish meaningful correspondences between virtual and real scenarios.
The core challenge lies in the inherent discrepancies between simulated inputs and real-world data, hindering cross-domain knowledge transfer, as traditional methods fail to establish meaningful correspondences between virtual and real scenarios.
Multimodal large language models (MLLMs), leveraging their robust reasoning~\cite{liu2024visual,zhong2025p3nav} and generalization capabilities~\cite{li2024llava,huang2024drivemm,yan2024robomm}, provide new opportunities for effectively merging different domain data, as demonstrated by the comparison of LLaVA-OneVision~\cite{li2024llava} with VAD in Fig.~\ref{fig:intro_img}.
Despite advancements, the misalignment in the geometric space between simulated and real data continues to constrain model performance.
Consequently, the critical research question remains unresolved:
% Consequently, the critical research question remains unresolved: How can MLLMs be systematically engineered to extract and transfer actionable driving knowledge from synthetic data while mitigating domain shift effects, thereby achieving substantial performance enhancements in real-world autonomous driving scenarios? 
\textit{How can MLLMs effectively leverage synthetic data to enhance real-world autonomous driving performance?}

To address the above two questions, this paper proposes RoboTron-Sim, a multimodal large language model framework designed to bridge the Sim2Real gap by learning actionable driving knowledge from simulated hard cases. 
First, we implement a data stratification strategy that categorizes real-world AD data and generates targeted synthetic scenarios through automated simulation pipelines. 
This approach specifically augments underrepresented safety-critical cases (\eg~long-tail scenarios, hard-to-drive scenarios), effectively mitigating the imbalance distribution problem. 
Second, we re-engineer the multimodal input schema with driving-aware prompts that enable dual-domain awareness and geographical conditioning. 
By explicitly encoding data provenance (Simulation/Real-World) and location-specific traffic patterns, the model dynamically leverages LLM-embedded commonsense knowledge to adjust driving policies while maintaining domain invariance. 
% (e.g., urban vs. highway)
Third, we introduce an image-to-ego encoder that explicitly injects camera parameters through a lightweight MLP-based adapter. 
This geometric-aware module disentangles dataset-specific sensor configurations from driving policy learning, significantly improving cross-dataset generalization capabilities. 
As shown in Fig.~\ref{fig:intro_img}, extensive experiments on nuScenes demonstrate that~\name achieves $\sim$50\% improvement in hard-case success rates compared to baseline methods, while maintaining performance in routine scenarios. 
% Our framework establishes new state-of-the-art in safety-critical driving metrics through systematic integration of simulated hard cases and physics-grounded domain adaptation. 

To summarize, our contributions lie in three-fold:
\begin{itemize}
    \item To our knowledge, we present the first in-depth investigation of Sim2Real transfer limitations in MLLMs for AD, accompanied by a simple yet effective framework that strategically leverages synthetic hard-case data to enhance real-world driving competencies.
    \item We address the critical domain discrepancy between simulated and real-world data, including scenario-aware prompts that dynamically model data provenance and geographical context, and a geometry-aware image-to-ego encoder that disentangles sensor-specific parameters.
    \item Experiments on nuScenes demonstrate \name{}achieves SOTA planning performance, particularly showing 48.1\% improvement in L2 Distance and 45.8\% enhancement in Collision Rate for hard scenarios. 

    % \item Extensive experiments on nuScenes show that \name{} achieves SOTA planning results, with 48.1\% lower L2 Distance and 45.8\% fewer Collisions in challenging scenarios.
    
    % while maintaining XX.X% effectiveness on routine driving tasks.   with xxxx overall improvement
\end{itemize}
\section{Related Work}
\paragraph{Autonomous Driving Models.}
% PLANNING方法
The development of autonomous driving systems has been significantly advanced by end-to-end approaches that established fundamental architectural paradigms. UniAD~\cite{hu2023planning} introduces unified perception-prediction-planning coordination through joint training frameworks, while VAD~\cite{jiang2023vad} employs vectorized scene representation methodology to optimize model computational efficiency.
In recent years, MLLMs have demonstrated exceptional potential in improving end-to-end planning. RDA-Driver~\cite{huang2024making} improves decision precision via reasoning-decision alignment and optimized CoT structures for interpretability. Senna~\cite{jiang2024senna} and DRIVEVLM~\cite{tian2024drivevlm} combine MLLMs with end-to-end models, leveraging the spatial perception capabilities of end-to-end models and the generalization ability of MLLMs to enhance planning.

% OmniDrive\cite{wang2024omnidrive} addresses 3D perception gaps with its Q-Former3D module bridging 2D-3D feature spaces. 
\vspace{-4mm}
\paragraph{Simulation Platforms for Autonomous Driving.}
% 合成仿真数据方法
Several open-source simulation platforms have been developed to facilitate autonomous driving research. 
CARLA~\cite{dosovitskiy2017carla} provides a modular urban driving simulator with configurable sensors (RGB/depth/semantic segmentation) and environmental conditions, enabling comparative evaluation of modular pipelines, imitation learning, and RL-based approaches. 
AirSim~\cite{shah2018airsim} delivers high-fidelity physical simulation through Unreal Engine, featuring photorealistic rendering with dynamic shadows/reflections and cross-platform APIs for perception-controller co-simulation. 
To tackle autonomous corner case challenges, OASIS SIM V3.0~\cite{yang2024oasis} leverages AI-powered traffic flow simulation with reinforcement learning and synthetic data to establish high-fidelity closed-loop environments, rigorously validating self-driving systems' rare scenario handling capabilities.

% ==== 数据分类-表格 ====
\begin{table*}[t]
\small
\centering
\addtolength{\tabcolsep}{0pt}
\begin{tabular}{l|cc|cc|cc}  % 使用 tabular 替代 tabularx
\toprule

Scenario & Day & Night & Sunny & Rainy & Straight & Turn \\
 
\midrule

Real Scenario & 24745 (87.97\%) & 3385 (12.03\%) & 22548 (80.16\%) & 5582 (19.84\%) & 24996 (88.86\%) & 3134 (11.14\%) \\
Simulated Scenario & 27891 (58.65\%) & 19662 (41.35\%) & 23010 (48.38\%) & 24543 (51.61\%) & 22076 (46.42\%) & 25477 (53.58\%) \\

\bottomrule
\end{tabular}
\vspace{-2mm}
\caption{Comparison of the volume of data from different sources across various scenarios.}
\label{tab:data class}
\end{table*}

\begin{figure*}[t]
    \centering
    \includegraphics[width=1\linewidth]{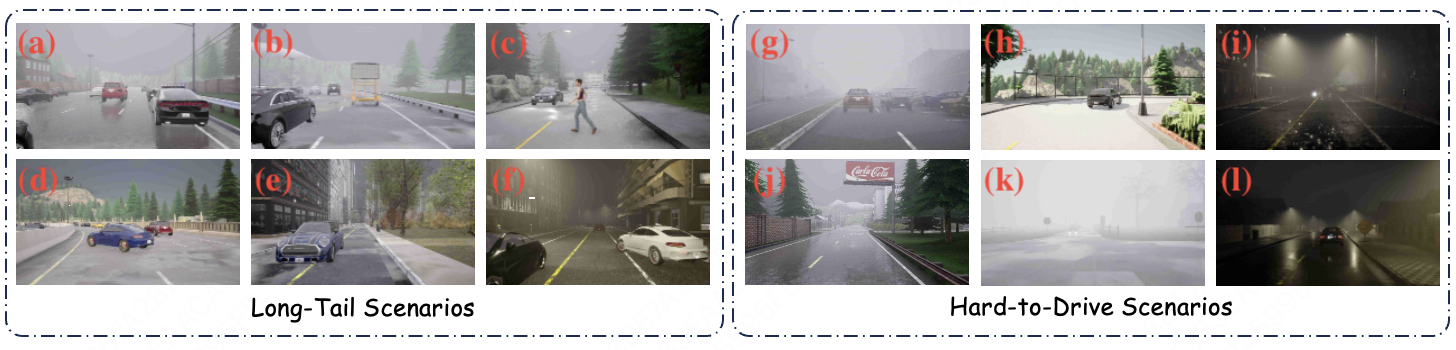}
    \vspace{-8mm}
    \caption{Visualization of long-tail and H2D scenarios in HASS. (a) Temporary Parking Ahead, (b) Roadwork Ahead, (c) Jaywalking Pedestrians, (d) Lane Invasion, (e) Opposing Lane Encroachment, (f) Parked vehicle Activation, (g)(j)(k) Rainy, (h) Turn, (i)(l) Night.}
    \vspace{-4mm}
    \label{fig:case_visual}
\end{figure*}

\begin{figure}[t] 
    \centering
    \includegraphics[width=0.48\textwidth]{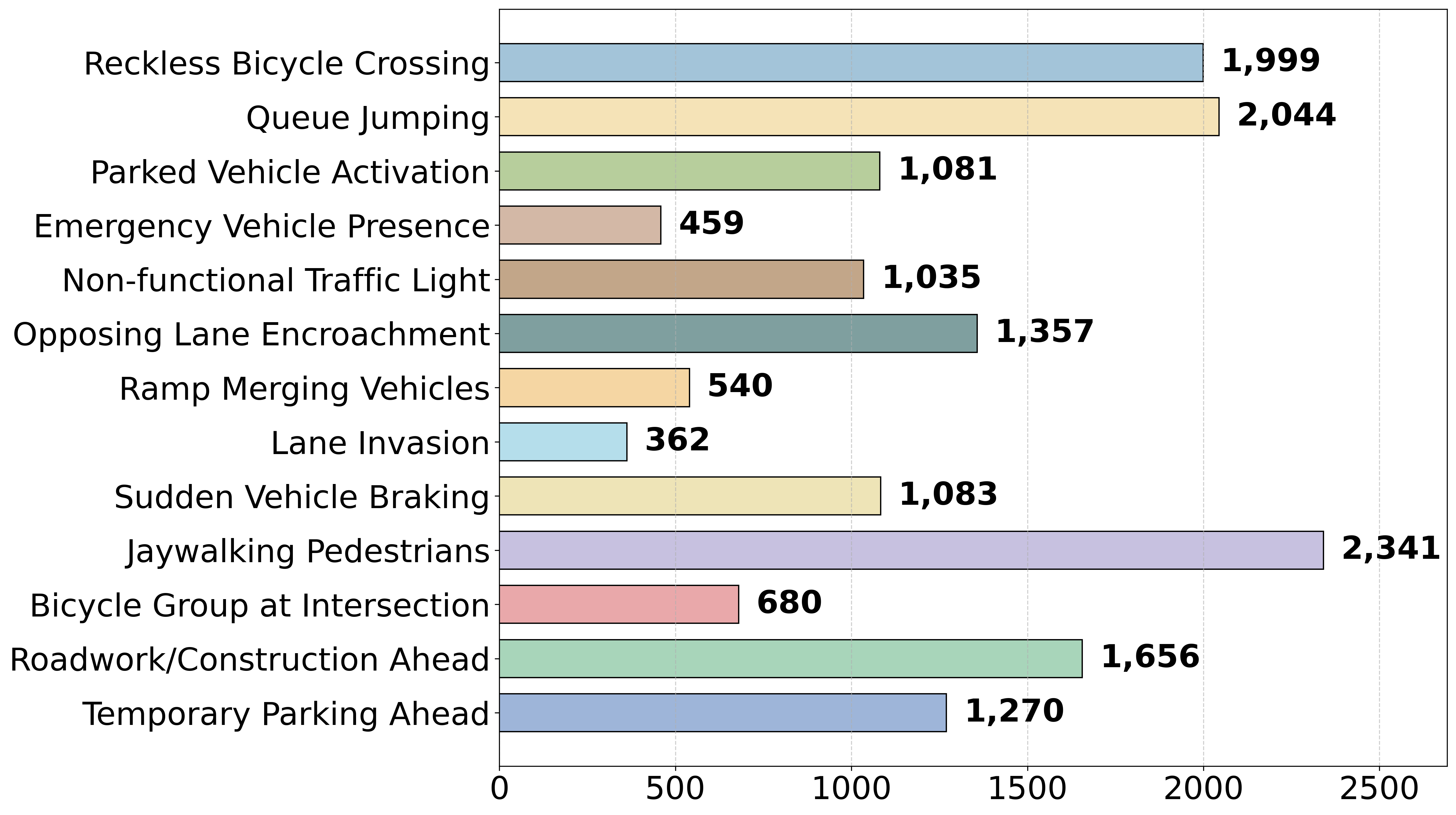} % 使图片适应单栏
    \vspace{-5mm}
    \caption{Types and sample counts of long-tail scenarios.}
    \vspace{-4mm}
    \label{fig:corner_cases}
\end{figure}

\section{Methodology}
In this section, our goal is to utilize simulated data to enhance the driving performance of the models in challenging and crucial scenarios, which are difficult for driving and data collection in real world.
To this end, we first create a simulated dataset Hard-case Augmented Synthetic Scenarios (HASS) that includes a variety of challenging and crucial scenarios in Sec.~\ref{sec:data_curation}. 
Subsequently, we introduce two dedicated technologies, Scenario-aware Prompt Engineering (SPE) and Image-to-Ego Encoder (I2E Encoder), designed to assist the MLLM baseline in effectively leveraging simulated data to enhance its performance in Sec.~\ref{sec:model}.
% we propose two dedicated technologies, prompt engineering and 3D position tokenizer, to effectively utilize the simulated data and enhance model performance in Sec.~\ref{sec:model}.
% Subsequently, we apply an MLLM-based approach to effectively leverage simulated data for enhancing the model's capability to address critical scenarios in Sec.~\ref{sec:model}.

\begin{figure*}[t]
    \centering
    \includegraphics[width=1\linewidth]{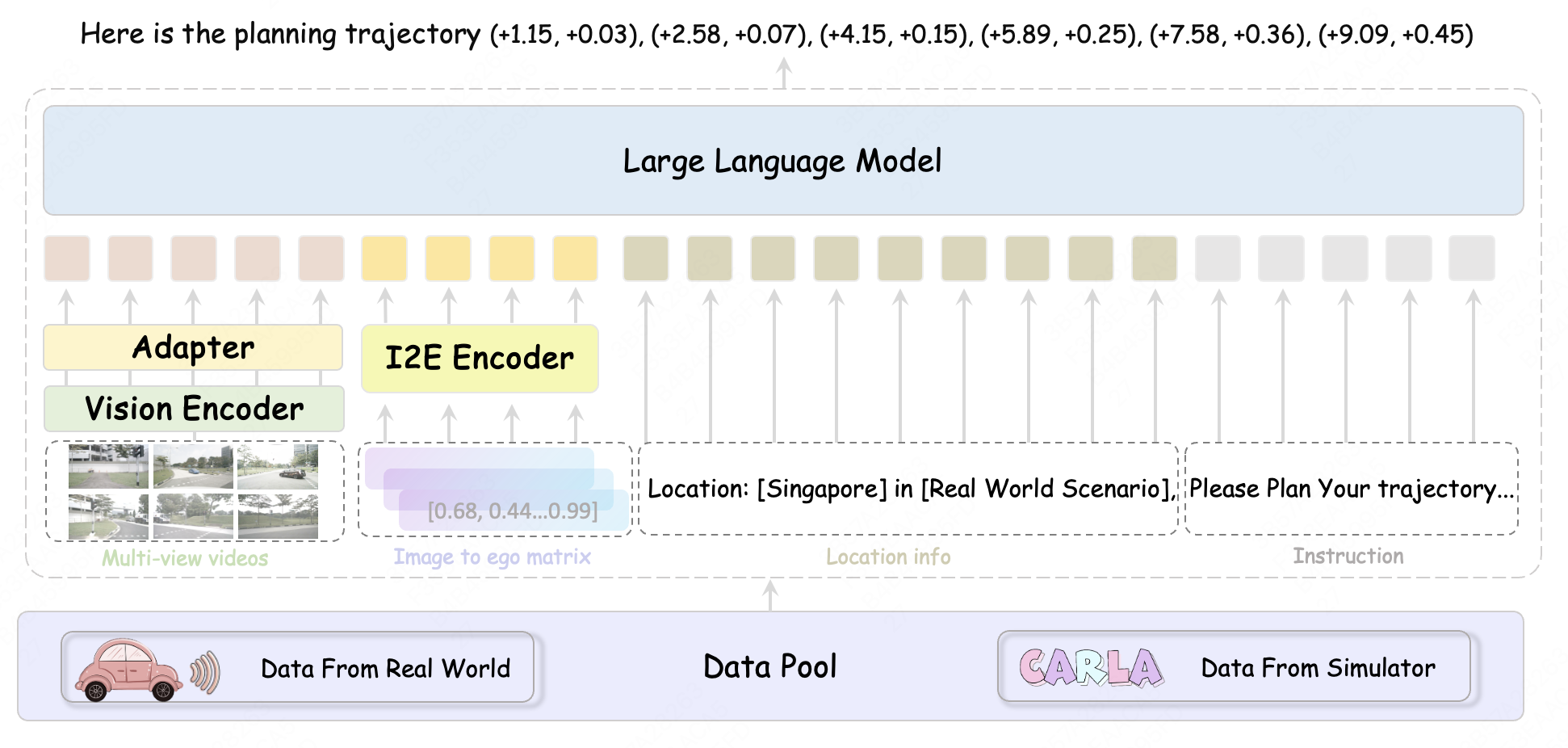}
    \vspace{-8mm}
    \caption{The overall framework of our proposed end-to-end autonomous driving system. The framework leverages simulation data to enhance performance in real-world scenarios, by integrating videos, 3D transformation, location, data source, and instruction information.}
    \vspace{-4mm}
    \label{fig:pipline}
\end{figure*}

\subsection{Data Curation}
\label{sec:data_curation}
% TODO: 可以合成各总hard case，可以利用专家信息生成专家轨迹
The effectiveness of synthetic data hinges on its ability to mirror real-world complexity while strategically addressing data scarcity in edge cases. We present a systematic framework for generating scenario-specific synthetic data in CARLA~\cite{dosovitskiy2017carla}, structured through three key dimensions: environmental diversity, agent behavior orchestration, and sensor-realistic multimodality.
\subsubsection{Scenario-Aware Data Classification}

To generate simulated data tailored to address the real-world challenging scenarios, we categorize the driving scenarios based on real-world data availability and criticality.
\vspace{-4mm}
\paragraph{Common Scenarios.}
Common scenarios are those frequently encountered in autonomous driving environments. Our scenario taxonomy systematically organizes the common driving scenarios into two distinct classes: Easy-to-Drive (E2D) and Hard-to-Drive (H2D) based on their driving difficulty. E2D scenarios refer to the uncomplicated common cases, such as daylight lane maintenance and uncongested traffic flows, which serve as foundational references for model calibration. In contrast, H2D refers to scenarios in which driving becomes challenging due to factors like weather and lighting. Examples include night-time driving, fog-obscured, and heavy-rain conditions where optical sensors degrade significantly.
Due to different occurrence frequencies of E2D and H2D scenarios and human driving preferences, the existing dataset often exhibits an imbalance between these two situations, e.g., the ratio of data volume between daytime and nighttime is approximately 7:1 in nuScenes. Nevertheless, \emph{H2D scenarios often require more data for training.} This motivates us to synthesize the H2D data to balance and enhance the driving performance in H2D scenarios.

% Our scenario taxonomy systematically organizes common driving scenarios into two distinct classes: Easy-to-Drive (E2D) and Hard-to-Drive (H2D) based on their driving difficulty. E2D scenarios represent frequently observed cases with abundant real-world data, such as daylight lane maintenance and uncongested traffic flows, which serve as foundational references for model calibration. In contrast, H2D scenarios, though equally common in occurrence, face systematic capture limitations due to environmental or sensor constraints. Examples include night-time driving, fog-obscured, and heavy-rain conditions where optical sensors degrade significantly. This dichotomy highlights the need for synthetic data to compensate for the underrepresentation of H2D scenarios in real datasets, even when their natural frequency matches E2D cases.
\vspace{-4mm}
\paragraph{Long-Tail Scenarios.}
% Long-tail scenarios encompass events that occur infrequently in natural driving environments yet exhibit high diversity and complexity. These scenarios often involve rare combinations of environmental factors, atypical agent behaviors, or transient conditions that challenge perception and decision-making systems. Examples include partially obscured traffic signs due to vegetation overgrowth, anomalous vehicle configurations (e.g., trucks carrying irregularly shaped cargo), and transient weather phenomena like localized fog patches or sun glare at low solar angles. Such scenarios defy conventional statistical modeling due to their combinatorial explosion of parameters. While individual events may have low occurrence probabilities, their cumulative likelihood becomes non-negligible across operational domains. The critical challenge lies not merely in their scarcity but in their inherent unpredictability, as these scenarios often violate implicit assumptions of spatial/temporal coherence embedded in perception models.

Long-tail scenarios encompass events that occur infrequently in natural driving environments yet exhibit high diversity and complexity. These scenarios often involve rare combinations of environmental factors, atypical agent behaviors, or transient conditions that challenge perception and decision-making systems. Examples include near-collision events, such as a vehicle narrowly avoiding a pedestrian suddenly crossing the road, abrupt pedestrian appearances during sharp turns, or unexpected vehicle maneuvers that nearly result in accidents. Other instances might involve complex interactions at intersections, such as a car running a red light or a pedestrian stepping into the roadway despite oncoming traffic. 
Such scenarios defy conventional statistical modeling due to their combinatorial explosion of parameters. 
\emph{While individual events may have low occurrence probabilities, they can lead to significant safety concerns.}
% their cumulative likelihood becomes non-negligible across operational domains. 
The critical challenge lies not merely in their scarcity but in their inherent unpredictability, as these scenarios often violate implicit assumptions of spatial/temporal coherence embedded in perception models.
By leveraging simulation data, which includes these specific long-tail scenarios, we can better train and evaluate the model's ability to handle rare but critical situations, ensuring robustness and safety in real-world applications.

\subsubsection{Data Collection}
\paragraph{Generation Details.}
Efficient data collection in simulation scenarios relies on teacher models equipped with global environmental awareness, which leverage privileged simulator-level information (e.g., precise poses of agents, intent semantics, and full traffic signal states) to achieve robust driving performance. Unlike conventional rule-based systems constrained by limited generalization across complex scenarios, we employ Think2Drive~\cite{li2024think2drive}, a world model-driven reinforcement learning architecture as the core data generator. 
To enhance reproducibility and lower the entry barrier for community-driven research in end-to-end AD, we implement a comprehensive sensor suite mirroring the nuScenes benchmark configuration, including six 900×1600 resolution cameras providing 360° coverage with overlapping fields of view. 

To systematically address both routine driving patterns and challenging long-tail cases mentioned above, we procedurally develop a dataset, called Hard-case Augmented Synthetic Scenarios (HASS). These scenarios are divided into two main categories: routine maneuvers (i.e., E2D and H2D) and long-tail scenarios. Several visualization examples of the H2D and long-tail scenarios are presented in Figure~\ref{fig:case_visual}. The long-tail scenarios are further classified into 13 high-risk edge case categories (e.g., pedestrian jaywalking, sudden vehicle cut-ins, and near-collision events), as visualized in Figure~\ref{fig:corner_cases}. These long-tail scenarios are designed to capture rare but critical driving scenarios that are often underrepresented in real-world datasets.

In addition to scenario diversity, HASS achieves balanced environmental conditions compared to traditional human-collected driving records. As shown in Table~\ref{tab:data class}, HASS shows a balanced distribution of environmental factors, including 58.65\% daytime and 41.35\% nighttime scenarios, as well as 48.38\% sunny and 51.61\% rainy conditions. Besides, HASS prioritizes interaction complexity, with 53.58\% of scenarios that involve turn-dominated maneuvers, a great increase compared to 11.14\% in real-world data. This deliberate design ensures that HASS not only addresses the limitations of real-world data but also provides a robust foundation for training and evaluating end-to-end AD systems in diverse and challenging conditions.

\vspace{-2mm}
% \subsubsection{Data Alignment}
\paragraph{Data Alignment.}
Real-world scenario (e.g., nuScenes) mainly adopts a right-handed coordinate system (X: forward, Y: left, Z: upward), while CARLA defaults to a left-handed coordinate system (X: forward, Y: right, Z: upward). This discrepancy involves both axial directional conflicts and spatial offsets in sensor positioning: in nuScenes, the vehicle coordinate origin is located at the center of the roof, while in CARLA, the vehicle coordinate origin is situated in the contact plane of the wheel. These conflicting axis definitions and origin offsets would otherwise cause cross-domain data fusion failures. To bridge the gap between simulated and real-world data distributions, we transform the coordinate system of the simulation data to a right-handed coordinate system and establish the vehicle's coordinate system origin as the center of the roof.

\subsection{Model}
\label{sec:model}
Our early efforts in tackling the planning task involved experimenting with traditional models like VAD~\cite{jiang2023vad}, trained on a mix of simulated and real-world data. However, as indicated by our experiments, the simulated data only yield very limited improvement
(reducing L2 distance by $\sim$1\% for the HD scenario).

We analyzed that the primary limitation stemmed from traditional models struggling to generalize effectively across diverse and dynamic environments, largely due to the substantial domain gap between simulation and reality. 
Recently, MLLMs have demonstrated remarkable generalization performance in various visual tasks.
This led us to explore the potential of MLLMs for addressing this task.

\begin{table}[t]
\begin{tcolorbox}[colback=white!95,
                  colframe=black,
                  width=\linewidth,
                  arc=1mm, auto outer arc,
                  boxrule=0.5pt,
                 ]
% \footnotesize
\small
\vspace{-1mm}
% \vspace{2mm}
\textbf{\textbf{Input:}} 

% Add a rounded rectangle around the six images
% \begin{tcolorbox}[colback=white, % Background color
%                   colframe=gray, % Border color
%                   arc=1mm, % Corner radius
%                   boxrule=0.5pt, % Border thickness
%                   width=\linewidth, % Full width
%                   left=0pt,right=0pt,top=0pt,bottom=0pt, % Remove padding
%                   ]
% \begin{center}
% % \includegraphics[width=0.32\linewidth]{1_bus.png}
% % \includegraphics[width=0.32\linewidth]{2_bus.png}
% % \includegraphics[width=0.32\linewidth]{3_bus.png}

% % \includegraphics[width=0.32\linewidth]{4_bus.png}
% % \includegraphics[width=0.32\linewidth]{5_bus.png}
% % \includegraphics[width=0.32\linewidth]{6_bus.png}

% \end{center}
% \end{tcolorbox}

\begin{tcolorbox}[colback=white, % Background color
                  colframe=gray, % Border color
                  arc=1mm, % Corner radius
                  boxrule=0.5pt, % Border thickness
                  width=\linewidth, % Full width
                  left=0pt,right=0pt,top=0pt,bottom=0pt, % Remove padding
                  ]
\begin{center}
\includegraphics[width=\linewidth]{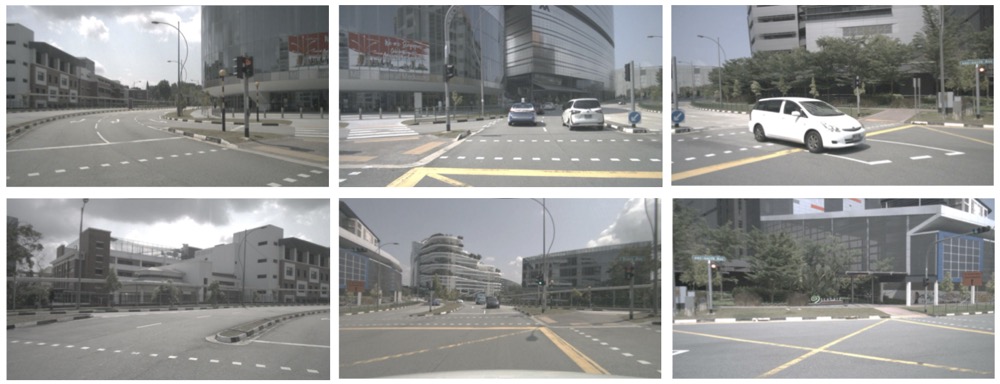}
\end{center}
\end{tcolorbox}

\textbf{\textbf{Question:}}
\vspace{1mm}

1: \textcolor{Red}{\textless video\textgreater}\;2: \textcolor{Red}{\textless video\textgreater} ... 6: \textcolor{Red}{\textless video\textgreater}. These 6 videos are the \textcolor{Purple}{front view}, \textcolor{Purple}{front left view}, \textcolor{Purple}{front right view}, \textcolor{Purple}{back view}, \textcolor{Purple}{back left view}, \textcolor{Purple}{back right view} of the ego vehicle. You need to \textbf{make a left turn at the upcoming intersection}, please provide the planning trajectory for the ego car.

\vspace{1mm}
\textbf{\textbf{Answer:}} 

Here is the planning trajectory (+4.96, +0.12), (+8.93, +0.48), (+12.62, +1.03), (+16.27, +1.78), (+19.67, +2.68), (+22.94, +3.70).

\end{tcolorbox}
\vspace{-4mm} 
\caption{The perspective-aware prompt for multi-view inputs.}
\vspace{-6mm}
\label{tab:prompt}
\end{table}

\subsubsection{MLLM Baseline}

% \begin{figure}
%     \centering
%     \includegraphics[width=1\linewidth]{image_pip2.png}
%     \caption{Enter Caption}
%     \label{fig:enter-label}
% \end{figure}

% \begin{figure}
%     \centering
%     \includegraphics[width=1\linewidth]{image_pip.png}
%     \caption{Enter Caption}
%     \label{fig:enter-label}
% \end{figure}

The proposed framework, as illustrated in Figure \ref{fig:pipline}, employs a multimodal architecture. It consists of three main components: a visual feature extractor, a feature adapter, and an LLM backbone.  
The visual feature extractor processes raw videos from multiple camera views over time, encoding them into a compact spatiotemporal representation, reducing dimensionality while preserving critical details for downstream tasks.
Furthermore, the framework employs a two-layer multilayer perceptron (MLP) to project the extracted visual features into the language model's embedding space, establishing dimensional compatibility between the visual representations and linguistic tokens while preserving semantic consistency across modalities.
Consequently, the LLM decoder integrates the processed visual features with tokenized text input, predicting textual outputs in an autoregressive manner. By leveraging both multimodal context and previously generated tokens, coherent and contextually relevant responses emerge.
\vspace{-3ex}
\paragraph{MLLM for planning.}
Building on the foundational MLLM architecture, we tailor the framework for end-to-end planning by enabling it to generate actionable trajectories from multimodal inputs. As shown in Tabel ~\ref{tab:prompt}, the model processes visual data captured from six cameras over five consecutive frames, alongside high-level command instructions (e.g., “make a left turn at the upcoming intersection,” “move forward”), ensuring a comprehensive action understanding. The visual feature extractor encodes spatiotemporal dynamics, while tokenized command instructions are fed into the LLM backbone alongside the projected visual features. To further enhance the model’s awareness of the ego vehicle’s state, we introduce velocity supervision. Consequently, the model produces two types of outputs: predicted trajectory points for the upcoming moments and anticipated vehicle speeds over the same timeframe.

% {
% \begin{table}[t]
% \begin{tcolorbox}[colback=white!95,
%                   colframe=black,
%                   width=\linewidth,
%                   arc=1mm, auto outer arc,
%                   boxrule=0.5pt,
%                  ]
% % \footnotesize
% \small
% \vspace{-1mm}
% % \vspace{2mm}
% \textbf{\textbf{Question:}} 

% \vspace{1mm}
% 1: \textcolor{Red}{\textless video\textgreater}\;2: \textcolor{Red}{\textless video\textgreater} ... 6: \textcolor{Red}{\textless video\textgreater}. These 6 videos are the \textcolor{Purple}{front view}, \textcolor{Purple}{front left view}, \textcolor{Purple}{front right view}, \textcolor{Purple}{back view}, \textcolor{Purple}{back left view}, \textcolor{Purple}{back right view} of the ego vehicle. You need to \textbf{make a left turn at the upcoming intersection}, please provide the planning trajectory for the ego car.
% \vspace{1mm}
% \textbf{\textbf{Answer:}} 

% Here is the planning trajectory [PT, (+4.96, +0.12), (+8.93, +0.48), (+12.62, +1.03), (+16.27, +1.78), (+19.67, +2.68), (+22.94, +3.70)]

% \end{tcolorbox}
% \vspace{-2mm}
% % \caption{Comparison of different prompts. \textbf{Top}: single-image prompt. \textbf{Bottom}: perspective-aware prompt.}
% \caption{The perspective-aware prompt for multi-view inputs.}
% \vspace{-6mm}
% \label{tab:prompt}
% \end{table}
% }

% 合并的表格
\begin{table*}[h]
\centering
\begin{tabularx}{1.0\textwidth}{l|*{4}{X}|*{4}{X}|*{4}{X}}
\toprule
\multirow{2}{*}{\textbf{Model}} &
\multicolumn{4}{c|}{\textbf{L2(m)}} & \multicolumn{4}{c|}{\textbf{Collision(\%)}} & \multicolumn{4}{c}{\textbf{Boundary(\%)}} \\
\cmidrule(r){2-5} \cmidrule(lr){6-9} \cmidrule(l){10-13}
& 1s & 2s & 3s & avg & 1s & 2s & 3s & avg & 1s & 2s & 3s & avg \\
\midrule
\multicolumn{13}{c}{\textbf{Without Ego Pose as Input}} \\
\midrule
UniAD\cite{hu2023planning} & 0.59 & 1.01 & 1.48 & 1.03 & 0.16 & 0.51 & 1.64 & 0.77 & \textbf{0.35} & \textbf{1.46} & \textbf{3.99} & \textbf{1.93} \\
VAD-Base\cite{jiang2023vad} & 0.69 & 1.22 & 1.83 & 1.25 & 0.06 & 0.68 & 2.52 & 1.09 & 1.02 & 3.44 & 7.00 & 3.82 \\
BEV-Planner\cite{li2024ego} & 0.27 & 0.54 & \textbf{0.90} & 0.57 & 0.10 & 0.37 & 1.30 & 0.59 & 0.78 &  3.79 & 8.22 & 4.26 \\
\midrule
LLaVA\cite{liu2023visual} & 1.04 & 1.74 & 2.57 & 1.79 & 0.58 & 1.17 & 1.74 & 1.16 & - & - & - & - \\
Vicuna\cite{chiang2023vicuna} & 1.06 & 1.80 & 2.54 & 1.80 & 0.60 & 1.21 & 1.78 & 1.20 & - & - & - & - \\
Merlin\cite{yu2024merlin} & 1.03 & 1.71 & 2.40 & 1.71 & 0.48 & 1.05 & 1.77 & 1.10 & - & - & - & - \\
OmniDrive\cite{wang2024omnidrive} & 0.40 & 0.80 & 1.32 & 0.84 & \textbf{0.04} & 0.46 & 2.32 & 0.94 & 0.93 & 3.65 & 8.28 & 4.29 \\
\name{} & \textbf{0.22} & \textbf{0.53} & 0.93 & \textbf{0.56} & 0.11 &\textbf{ 0.35} & \textbf{1.27} & \textbf{0.58} & 0.55 & 1.97 & 5.57 & 3.02 \\
\midrule
\multicolumn{13}{c}{\textbf{With Ego Pose as Input}} \\
\midrule
UniAD\cite{hu2023planning} & 0.20 & 0.42 & 0.75 & 0.46 & 0.02 & 0.25 & 0.84 & 0.37 & \textbf{0.20} & \textbf{1.33} & \textbf{3.24} & \textbf{1.59} \\
VAD-Base\cite{jiang2023vad} & 0.17 & 0.34 & 0.60 & 0.37 & 0.04 & 0.27 & 0.67 & 0.33 & 0.21 & 2.13 & 5.06 & 2.47 \\
Ego-MLP\cite{zhai2023rethinking} & 0.15 & 0.32 & 0.59 & 0.35 & 0.00 & 0.27 & 0.85 & 0.37 & 0.27 & 2.52 & 6.60 & 2.93 \\
BEV-Planner\cite{li2024ego} & 0.16 & 0.32 & 0.57 & 0.35 & 0.00 & 0.29 & 0.73 & 0.34 & 0.35 & 2.62 & 6.51 & 3.16 \\
\midrule
EMMA\cite{hwang2024emma} & 0.14 & 0.29 & 0.54 & 0.32 & - & - & - & - & - & - & - & - \\
OmniDrive\cite{wang2024omnidrive} & 0.14 & 0.29 & 0.55 & 0.33 & 0.00 & 0.13 & 0.78 & 0.30 & 0.56 & 2.48 & 5.96 & 3.00 \\
% Ours & 0.13 & 0.28 & 0.55 & 0.32 & 0.00 & 0.01 & 0.68 & 0.26 & 0.53 & 2.63 & 6.62 & 3.26 \\
% Ours & 0.13 & 0.28 & 0.52 & 0.31 & 0.00 & 0.09 & 0.82 & 0.31 & 0.41 & 1.97 & 5.57 & 2.65 \\
\name{} & \textbf{0.10} & \textbf{0.20} & \textbf{0.39} & \textbf{0.23} & \textbf{0.00} & \textbf{0.11} & \textbf{0.66} & \textbf{0.26} & 0.39 & 2.07 & 5.41 & 2.62 \\
\bottomrule
\end{tabularx}
\vspace{-2mm}
\caption{Comparison on the open-loop planning on nuScenes dataset with and without ego pose as input. We report the L2 distance (m), collision rate (\%), and boundary violation rate (\%) at 1s, 2s, and 3s time horizons, along with their averages.}
\vspace{-3mm}
\label{tab:combined_table}
\end{table*}

\subsubsection{Enhancing Driving with Simulated Data}
% While directly mixing simulated data with real-world data during training 
We attempted to train the designed baseline by directly mixing the real-world data with simulated data. Although this approach has shown some improvements (reducing L2 distance from 1.03m to 0.91m), we identified that this approach alone is insufficient to fully adapt to the inherent differences between data sources, limiting the model's ability to achieve robust generalization.
To better align and leverage the complementary strengths of simulated data, we introduce two key designs aimed at enhancing the MLLM baseline to transfer diverse simulated driving skills to real-world. The resultant model is named RoboTron-Sim. 
% To better align and leverage the complementary strengths of simulated and real-world data, we introduced two key designs aimed at enhancing the MLLM's understanding of diverse driving scenarios.

\vspace{-4mm}
\paragraph{Scenario-Specific Contextual Grounding.} 
To address the domain discrepancy between simulated and real-world environments, we implement contextual priming through Scenario-aware Prompt Engineering (SPE). Specifically, each input sequence is augmented with an environmental descriptor formatted as: \textit{``You are driving in [City Name] under [Simulation/Real-World] scenario."} This structured prompt serves dual purposes:
\begin{itemize}
    \item Domain Awareness: Explicitly informs the model about the data characteristics (e.g., sensor noise levels) through categorical labeling of simulation/real-world contexts.
    \item Geographical Conditioning: Embeds location-specific priors (e.g., traffic rules) via city name specification, enabling adaptive processing of regional driving patterns.
\end{itemize}
For instance, the prompt \textit{``You are driving in Town13 under simulation scenario"} activates the model's knowledge of both the city's unique driving habits (left-hand or right-hand driving) and traffic rules. This added context helps the MLLM baseline adjust its decision-making, ignoring unrealistic details from simulations when working with real-world data, while making use of useful patterns learned from synthetic scenarios to improve planning stability.

\vspace{-3mm}
\paragraph{Geometry-Aware Visual Alignment.}
Due to variations in vehicles and cameras, together with different sensor installation positions in simulated and real-world scenarios, the intrinsic and extrinsic parameters of the camera typically differ between the two systems.
This discrepancy creates a critical cross-domain gap, leading to degraded performance when transferring between simulation and reality. 
To enhance adaptability to different camera parameters, we introduce an Image-to-Ego Encoder (I2E Encoder) to explicitly incorporate geometric transformations into the input processing, enhancing the visual features with 3D spatial information, as shown in Figure~\ref{fig:pipline}.
Specifically, we first compute the image-to-ego transformation matrix using the camera's intrinsic and extrinsic parameters for each view, ensuring a consistent spatial representation across different viewpoints. Subsequently, we employ the I2E Encoder, implemented as a two-layer MLP, to integrate this geometric information into the MLLM baseline.
This encoder maps the transformation matrix into an embedding space that captures the spatial context of each camera view. The resulting embeddings are then concatenated with the tokenized text input, allowing the model to incorporate spatial reasoning directly into its decision-making process.

\vspace{-2mm}
\section{Experiment}
\subsection{Experimental Setting}
\subsubsection{Dataset}
\name{}is trained using a hybrid data strategy combining:
\begin{itemize}
    \item \textbf{Real-world Data}: 28,130 samples from nuScenes\cite{caesar2020nuscenes}.
    
    \item \textbf{Simulated Data}: 47,553 purpose-built samples generated in CARLA simulator \cite{dosovitskiy2017carla}.

    % , designed to address the inherent imbalance in real-world data distribution. 
    
    % While the dataset covers a broad range of driving situations, it places particular emphasis on addressing challenging cases, including H2D scenarios and Long-Tail scenarios.
    
    % \item \textbf{Test Set}: 6,019 samples from nuScenes validation split, ensuring evaluation consistency with real-world benchmarks
\end{itemize}

\noindent For evaluation, we use nuScenes validation set. Please refer to the supplementary material for more details.
\vspace{-2mm}
\subsubsection{Evaluation Metrics}
Following BEV-Planner~\cite{li2024ego}, we evaluate via L2 Distance, Collision Rate, and Boundary Violation Rate. Please refer to the supplementary material for more details.

% \begin{itemize}
%     \item \textbf{Trajectory Accuracy (L2 Distance)}:  
%     \begin{equation}
%         L2 = \frac{1}{T}\sum_{t=1}^{T} \| \hat{p}_t - p^{gt}_t \|_2
%     \end{equation}
%     where $\hat{p}_t$ and $p^{gt}_t$ denote the predicted and ground-truth positions at timestep $t$ over a $T=3s$ horizon.  

%     \item \textbf{Safety Metrics (Collision Rate)}:  
        
%         Computes the percentage of predicted trajectories that result in collisions with other agents or obstacles.  
%         \begin{equation}
%     Collision = \frac{1}{T} \sum_{t=1}^{T} \frac{N_{\text{collision}, t}}{N_{\text{total}, t}} \times 100\%
%         \end{equation}

%         where $N_{\text{collision}}$ is the number of predicted trajectories leading to collisions, and $N_{\text{total}}$ is the total number of evaluated trajectories at timestep $t$ over a $T=3s$ horizon.  

% \end{itemize}

% \begin{itemize}
%     \item \textbf{Boundary Violation Rate}:  
%     \begin{equation}
%         Boundary = \frac{1}{T} \sum_{t=1}^{T} \frac{N_{\text{violation}, t}}{N_{\text{total}, t}} \times 100\%
%     \end{equation}
%     where $N_{\text{violation}}$ counts trajectories exceeding road boundaries, and $N_{\text{total}}$ is the total evaluated trajectories at timestep $t$ over $T=3s$. Calculated by comparing ego segmentation masks with drivable area labels.
    
% \end{itemize}

\subsubsection{Training Details}
Our model architecture builds upon LLaVA-OneVision~\cite{li2024llava}, with continuous five-frame sequences from six camera views as input. The training was performed on 16 NVIDIA A100 GPUs for 3 epochs, taking approximately 20 hours when using nuScenes and HASS.

%  % ==== 消融实验-表格 ====
% \begin{table*}[t]
% \centering
% \addtolength{\tabcolsep}{10pt}
% \begin{tabular}{l|cc|ccc}  % 使用 tabular 替代 tabularx
% \toprule

% Data & SPE & I2E Encoder & L2(m) & Collision(\%) & Boundary(\%) \\
 
% \midrule

% nuScenes & \xmarkg & \xmarkg & 1.03 & 1.01 & 3.45   \\
% \midrule
% nuScenes + HASS & \xmarkg & \xmarkg & 0.91 & 0.94 & 3.22 \\
% nuScenes + HASS & \cmarkg & \xmarkg & 0.86 & 0.79 & 2.68 \\
% nuScenes + HASS & \cmarkg & \cmarkg & 0.56 & 0.58 & 3.02 \\

% \bottomrule
% \end{tabular}
% % \vspace{4pt}
% \caption{Ablation on different experimental input contexts.}
% \label{tab:traing_data_ablation}
% \end{table*}

 % \name{}outperforms existing approaches in both settings, demonstrating its ability to leverage multimodal inputs effectively and adapt to varying levels of available information.\name{}demonstrates robust performance across scenarios with and without ego pose data.

\begin{table*}[t]
\centering
\scalebox{0.8}{
\addtolength{\tabcolsep}{2pt}
\begin{tabular}{l|c|lll|lll}
\toprule
\multirow{2}{*}{Method} & \multirow{2}{*}{Data} & 
\multicolumn{3}{c|}{E2D} & 
\multicolumn{3}{c}{H2D} \\
\cmidrule(lr){3-5} \cmidrule(l){6-8}
& & Day & Straight & Sunny & Night & Turn & Rainy  \\
\midrule

\multirow{2}{*}{VAD} 
 & nuScenes  & 0.77 & 0.78 & 0.78 & 0.94 & 0.87 & 0.83 \\
 & nuScenes + HASS & 
 0.74$\color{red}({\downarrow{3.9\%}})$ & 
 0.78$\color{gray}({\downarrow{0.0\%}})$ & 
 0.75$\color{red}({\downarrow{3.8\%}})$ & 
 0.93$\color{red}({\downarrow{1.1\%}})$ & 
 0.87$\color{gray}({\downarrow{0.0\%}})$ & 
 0.82$\color{red}({\downarrow{1.2\%}})$ \\
\midrule

\multirow{2}{*}{MLLM} 
 & nuScenes & 1.00 & 1.01 & 1.00 & 1.33 & 1.13 & 1.15 \\
 & nuScenes + HASS & 
 0.88$\color{red}({\downarrow{12.0\%}})$ & 
 0.91$\color{red}({\downarrow{9.9\%}})$ & 
 0.92$\color{red}({\downarrow{8.0\%}})$ & 
 1.31$\color{red}({\downarrow{1.5\%}})$ & 
 1.02$\color{red}({\downarrow{9.7\%}})$ & 
 0.90$\color{red}({\downarrow{21.7\%}})$ \\
\midrule

\multirow{2}{*}{RoboTron-Sim} 
 & nuScenes & 0.59 & 0.59 & 0.64 & 1.40 & 1.32 & 1.15 \\
 & nuScenes + HASS & 
 0.54$\color{red}({\downarrow{8.5\%}})$ & 
 0.55$\color{red}({\downarrow{6.8\%}})$ & 
 0.56$\color{red}({\downarrow{12.5\%}})$ & 
 0.81$\color{red}({\downarrow{42.1\%}})$ & 
 0.64$\color{red}({\downarrow{51.5\%}})$ & 
 0.56$\color{red}({\downarrow{51.3\%}})$ \\
\bottomrule
\end{tabular}}
\vspace{-3mm}
\caption{L2 Performance gains of HASS on different models in E2D and H2D scenarios. 
Red arrows indicate improvement (\textcolor{red}{lower values are better}), gray denotes no significant change. ``MLLM" means the baseline obtained by finetuning LLaVA-OneVison on driving data.}
\vspace{-3mm}
\label{tab:traing_data_ablation3}
\end{table*}

\begin{figure}
    \centering
    \includegraphics[width=1\linewidth]{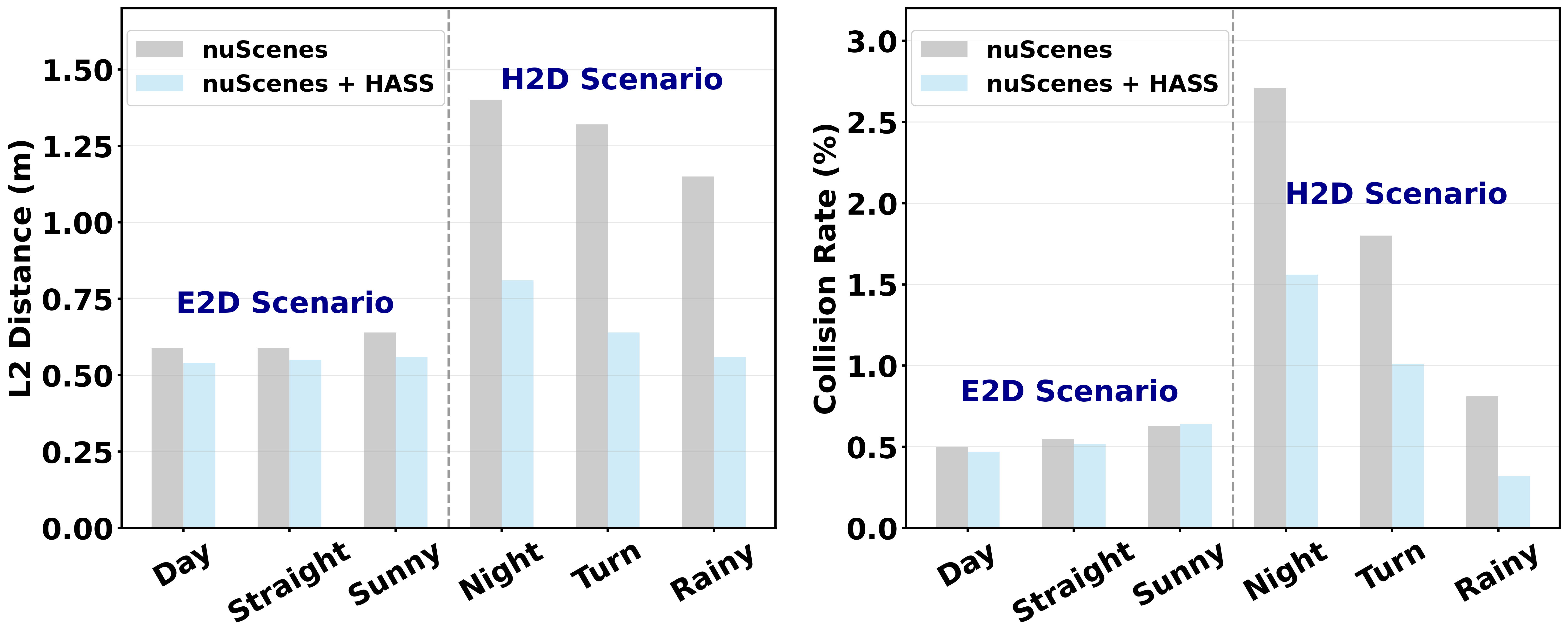}
    % \vspace*{0pt} 
    \vspace{-3mm}
    \caption{Comparison of \name{}on different datasets.}
    \vspace{-3mm}
    \label{fig:env_comp}
\end{figure}

\subsection{Main Results}
\paragraph{Open-loop Planning.}
To ensure a fair comparison of planning capabilities across different methods, we evaluate our method under two settings: (1) without ego pose as input and (2) with ego pose as input. As shown in Tab.~\ref{tab:combined_table}, when ego pose is excluded, the framework achieves state-of-the-art trajectory prediction accuracy with the lowest average L2 distance (0.56m) and collision rate (0.58\%) while maintaining a competitive boundary violation rate (3.02\% vs. 4.29\% in OmniDrive). Incorporating ego pose further enhances these metrics, yielding substantial improvements to 0.23m L2 distance and 0.26\% collision rate, achieving the SOTA. The boundary violation rate also decreases to 2.62\%, demonstrating consistent performance gains across all safety-critical metrics. This highlights RoboTron-Sim's adaptability to both sensor-limited and sensor-rich conditions while maintaining superior safety margins.
% \paragraph{Without Ego Pose Data.}
% When Ego Pose data is excluded, our method demonstrates superior performance in both trajectory prediction and collision avoidance. Specifically, our approach achieves the lowest average L2 distance (0.56m) and collision rate (0.58\%), outperforming state-of-the-art methods such as UniAD (1.03m, 0.77\%) and BEV-Planner (0.57m, 0.59\%). Notably, our framework maintains a competitive boundary violation rate (3.02\%), which is significantly lower than most baselines, including OmniDrive (4.29\%). These results highlight the robustness of our method in handling complex driving scenarios even without direct access to ego vehicle dynamics.
% \paragraph{With Ego Pose Data.}
% When Ego Pose data is included, our method further improves its performance, achieving the lowest average L2 distance (0.23 m) and collision rate (0.26\%). Our method outperforms strong baselines such as Ego-MLP (0.35 m, 0.37\%) and OmniDrive (0.33 m, 0.30\%) in terms of L2 distance and collision rate, respectively. Additionally, our boundary violation rate (2.62\%) remains competitive, further validating the effectiveness of our approach.

\vspace{-4mm}
\paragraph{Scenario-Specific Improvement.}
Here we evaluate the performance improvement of \name{}in each scenario. Specifically, we train the proposed model on 
% We evaluate HASS's capabilities in hard cases through training \name{}with 
two kinds of data: (1) nuScenes, and (2) nuScenes + HASS. 
The results presented in Fig.~\ref{fig:env_comp} expose striking performance disparities across environments. In E2D scenarios (Day + Straight + Sunny) where existing models already demonstrate mature capabilities, the performance ceiling leaves limited optimization space. Conversely, in H2D scenarios (Night + Turn + Rainy), the integration of HASS drives remarkable progress: the nighttime collision rate is reduced by \textbf{42.4\%} (from 2.71\% to 1.56\%), while turning maneuver precision shows \textbf{51.5\%} improvement in L2 metrics. It is worth noting that \name{}achieves \textbf{51.3\%} lower collision rates than the baseline approach. This flipped contrast underscores the value of HASS: while maintaining stability in routine scenarios, it primarily empowers breakthrough advancements in hard cases via synthetic scenario. 

\begin{figure}
    \centering
    \includegraphics[width=1\linewidth]{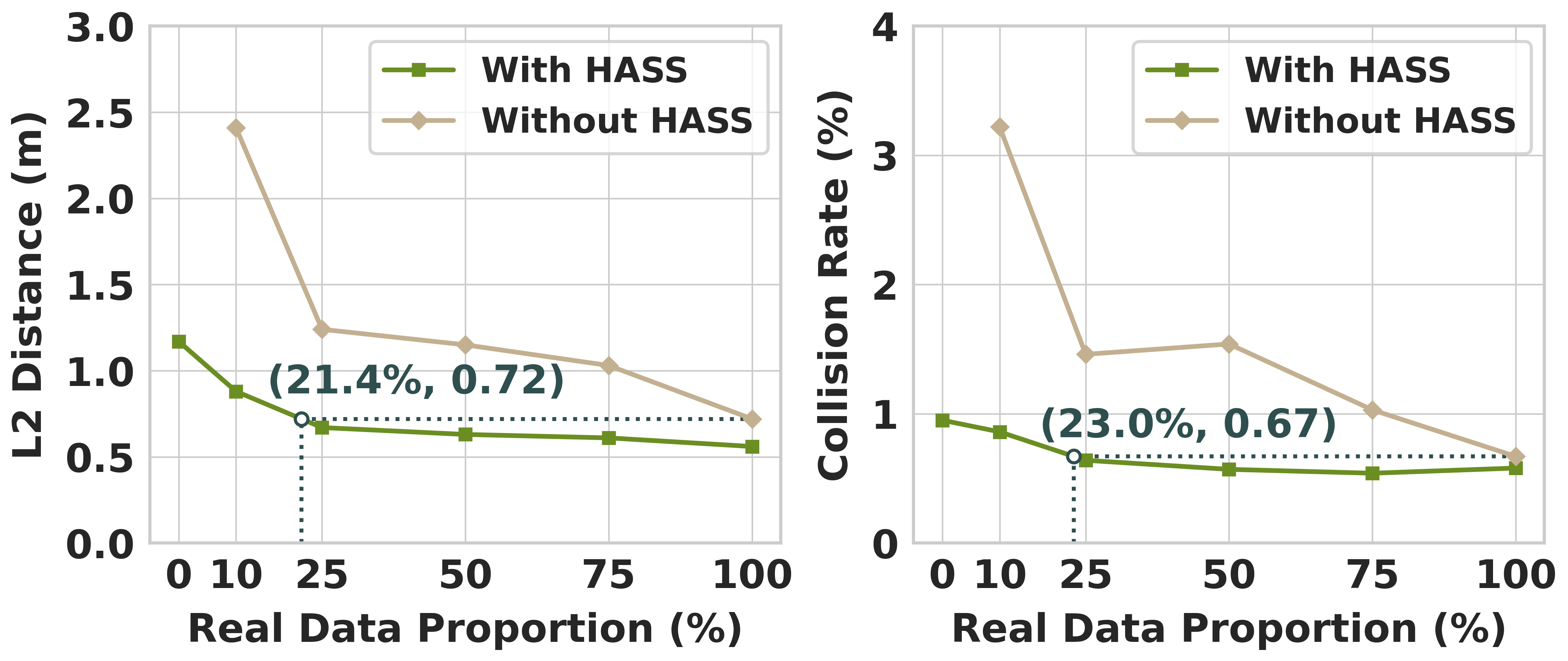}
    % \vspace*{-4pt} 
    \vspace{-3mm}
    \caption{
Performance comparison between methods \textbf{with} and \textbf{without} HASS under different ratios of nuScenes data.}
    \vspace{-3mm}
    \label{fig:sample_exp}
\end{figure}

\vspace{-4mm}
\paragraph{Data Robustness.}  
We conduct comparative experiments in multiple data regimes to verify the robustness of the data:
% establishing two settings: 
(1) different proportions of nuScenes data (from 100\% to 0\%), and (2) different proportions of nuScenes data + full HASS data.
As shown in Fig.~\ref{fig:sample_exp}, \name{}maintains stable performance even as the proportion of real-world data decreases from 100\% to 10\%. In contrast, purely real-data-based models experience severe performance degradation in limited-data conditions, with both L2 distance and collision rate increasing significantly as real samples diminish. By incorporating HASS, performance degradation is alleviated, ensuring minimal performance loss even when real data is scarce. Notably, we find that by combining simulated data with only around 20\% real samples, we can achieve the same level of performance as using 100\% real data. This highlights the crucial role of simulated data in compensating for real-world data limitations while preserving robust trajectory planning across diverse data distributions.

% \begin{table*}[t]
% \centering
% \addtolength{\tabcolsep}{2pt} % 适当减少列间距
% \begin{tabular}{l|c|ll|ll}
% \toprule
% \multirow{2}{*}{Method} & \multirow{2}{*}{Data} & 
% \multicolumn{2}{c|}{E2D} & 
% \multicolumn{2}{c}{H2D} \\
% \cmidrule(lr){3-4} \cmidrule(l){5-6}
% & & L2(m) & Collision(\%) & L2(m) & Collision(\%)  \\
% \midrule

% \multirow{2}{*}{VAD} 
%  & nuScenes & 0.78 & 0.26 & 0.88 & 0.35 \\
%  & nuScenes + HASS & 
%  0.76$\color{blue}({\downarrow{2.5\%}})$ & 
%  0.36$\color{red}({\uparrow{38.5\%}})$ & 
%  0.87$\color{blue}({\downarrow{1.1\%}})$ & 
%  0.43$\color{red}({\uparrow{22.9\%}})$ \\
% \midrule

% \multirow{2}{*}{MLLM} 
%  & nuScenes &  &  &  &  \\
%  & nuScenes + HASS &  &  &  &  \\

% \midrule

% \multirow{2}{*}{RoboTron-Sim} 
%  & nuScenes & 0.61 & 0.56 & 1.29 & 1.77 \\
%  & nuScenes + HASS & 
%  0.57$\color{blue}({\downarrow{6.6\%}})$ & 
%  0.54$\color{blue}({\downarrow{3.6\%}})$ & 
%  0.67$\color{blue}({\downarrow{48.1\%}})$ & 
%  0.96$\color{blue}({\downarrow{45.8\%}})$ \\
% \bottomrule
% \end{tabular}
% \vspace*{-4pt} 
% \caption{Performance gains of HASS across different models in E2D and H2D scenarios. To rigorously evaluate the model's inherent capability to comprehend dynamic environments without relying on ego-pose dependencies, we conducted ablation studies by removing ego-pose inputs from both MLLM and \name{}architectures.}
% \label{tab:traing_data_ablation3}
% \end{table*}

\begin{table*}[t]
\centering
\scalebox{0.9}{
\addtolength{\tabcolsep}{2pt}
\begin{tabular}{l|lll|lll}
\toprule
\multirow{2}{*}{Data} & 
\multicolumn{3}{c|}{E2D} & 
\multicolumn{3}{c}{H2D} \\
\cmidrule(lr){2-4} \cmidrule(l){5-7}
& Day & Straight & Sunny & Night & Turn & Rainy  \\
\midrule

nuScenes & 0.59 & 0.59 & 0.64 & 1.40 & 1.32 & 1.15 \\
\midrule

nuScenes + GASS & 
0.55$\color{red}({\downarrow{6.8\%}})$ & 
0.50$\color{red}({\downarrow{15.3\%}})$ & 
0.52$\color{red}({\downarrow{18.8\%}})$ & 
1.00$\color{red}({\downarrow{28.6\%}})$ & 
1.21$\color{red}({\downarrow{8.3\%}})$ & 
0.99$\color{red}({\downarrow{13.9\%}})$ \\
\midrule

nuScenes + HASS & 
0.54$\color{red}({\downarrow{8.5\%}})$ & 
0.55$\color{red}({\downarrow{6.8\%}})$ & 
0.56$\color{red}({\downarrow{12.5\%}})$ & 
0.81$\color{red}({\downarrow{42.1\%}})$ & 
0.64$\color{red}({\downarrow{51.5\%}})$ & 
0.56$\color{red}({\downarrow{51.3\%}})$ \\
\bottomrule
\end{tabular}}
\vspace{-2mm}
\caption{L2 Performance comparison of different datasets in E2D and H2D scenarios (\textcolor{red}{lower values are better}).}
\label{tab:sim_scenario_comparison}
\vspace{-3mm}
\end{table*}

% ==== 消融实验-表格 ====
\begin{table}[t]
\centering
\scalebox{0.90}{
\begin{minipage}{\textwidth} 
\small
\addtolength{\tabcolsep}{0pt}
\begin{tabular}{cc|ccc}  % 使用 tabular 替代 tabularx
\toprule

SPE & I2E Encoder & L2(m) & Collision(\%) & Boundary(\%) \\
 
\midrule

\xmarkg & \xmarkg & 0.91 & 0.94 & 3.22 \\
\cmarkg & \xmarkg & 0.86 & 0.79 & 2.68 \\
\cmarkg & \cmarkg & 0.56 & 0.58 & 3.02 \\

\bottomrule
\end{tabular}
\end{minipage}
}
% \vspace{4pt}
\vspace*{-4pt} 
\caption{Ablation study on model designs.}
\vspace{-4mm}
\label{tab:traing_data_ablation}
\end{table}

\subsection{Ablation Study}
\paragraph{Model Compatibility.}
% To systematically investigate model compatibility with simulated data augmentation, we conduct cross-architecture evaluations on L2 distance. 
% As evidenced in Table~\ref{tab:traing_data_ablation3}, VAD exhibits fundamental compatibility limitations, with marginal L2 reductions (↓2.5\% E2D, ↓1.1\% H2D) overshadowed by significantly increased collision rates (↑38.5\% E2D, ↑22.9\% H2D). Although MLLM demonstrates preliminary compatibility, showing gradual improvements, the gains remain constrained in the hard cases. In stark contrast, our \name{}achieves breakthrough enhancements (↓48.1\% L2, ↓45.8\% collision rate) in H2D case. This empowers knowledge transfer from synthetic domains while preserving real-world physical constraints, unlocking the model's untapped potential.

To investigate model compatibility with simulated data augmentation, we conduct cross-architecture evaluations on L2 distance across driving scenarios. As evidenced in Table~\ref{tab:traing_data_ablation3}, the effectiveness of HASS varies significantly across architectures:
VAD exhibits inherent incompatibility with HASS, showing marginal L2 improvements (3-4\% in daytime/sunny conditions) but showing negligible gains (below 1.5\%) in H2D scenarios. While MLLMs demonstrate preliminary adaptability with moderate L2 reductions (8-12\% in E2D), their gains decline in H2D scenarios, exposing limitations in modeling vehicle dynamics. In contrast, \name{}achieves paradigm-shifting enhancements through Sim2Real alignment, it reduces L2 distance by \textbf{51.5\%} in complex turns and \textbf{42-51\%} in night/rain conditions. This empowers knowledge transfer from synthetic domains while preserving real-world physical constraints, unlocking the model's untapped potential.

\vspace{-3mm}
\paragraph{Ablation on Model Designs.}
% \paragraph{Ablation on SPE.}
% The third row introduces SPE, which explicitly primes the model with domain-awareness regarding real-world vs. simulated data contexts. Enabling SPE further reduces the L2 distance from 0.91m to 0.86m and the collision rate from 0.94\% to 0.79\%, while also significantly decreasing boundary violations. This suggests that SPE effectively mitigates domain discrepancies by guiding the model to account for variations in driving habits, and environmental factors, thereby enhancing generalization across domains.
Tab.~\ref{tab:traing_data_ablation} (rows 1-2) shows SPE's enhancements: L2 is reduced by 5.5\%, collisions drops by 16.0\%, and most notably, boundary violations see a substantial decrease of 38.2\% (3.22\%→2.68\%), which proves SPE's ability to enhance the model's sensitivity to lane discipline and road geometry constraints.
% \paragraph{Ablation on I2E Encoder.}
The last two rows of Tab.~\ref{tab:traing_data_ablation} show that integrating I2E Encoder reduces L2 distance by 34.9\% and collision rate by 26.6\%. This demonstrates that explicit geometric grounding effectively bridges the Sim2Real domain gap by establishing camera-independent spatial representations, enabling consistent perception across synthetic and physical sensors.

\vspace{-3mm}
\paragraph{Effectiveness of HASS.}
To evaluate the impact of different data synthesis strategies, we conduct experiments using two kinds of synthetic data: (1) Hard-case Augmented Synthetic Scenarios (HASS), where we adjust the synthesis proportion to focus on hard cases, and (2) General Augmented Synthetic Scenarios (GASS), where the data is synthesized following the scene distribution in nuScenes. As shown in Tab.~\ref{tab:sim_scenario_comparison}, the method trained on GASS performs well across E2D scenarios but struggles in H2D conditions, such as night and turn scenarios, where higher L2 errors are observed (1.00m and 1.21m). In contrast, the model trained on HASS, which increases the representation of hard cases, achieves great improvements in H2D scenarios. Please refer to the supplementary material for more details.

\vspace{-1mm}
\subsection{Case Study}

% We visualize the predicted trajectory points from both the baseline and \name{}under ego-pose-free conditions, alongside the corresponding ground-truth trajectory points. As shown in Fig.~\ref{fig:case-a}, when confronted with an oncoming vehicle crossing lane markings and a stationary vehicle ahead in the ego lane, \textcolor{green}{RoboTron-Sim} aligns with the \textcolor{red}{ground truth} in executing a safety stop, while the \textcolor{yellow}{baseline} dangerously persists forward motion. This behavior reflects our model's improved risk awareness in complex interactions. \name{}also shows enhanced safety for sharp turning cases (Fig.~\ref{fig:case-c}) These visual cases corroborate the results in Fig.~\ref{fig:env_comp}, confirming the effectiveness of \name{}in safety-critical cases. 

\begin{figure}[t]
    \centering
    \begin{subfigure}[b]{0.23\textwidth}
        \includegraphics[width=\textwidth]{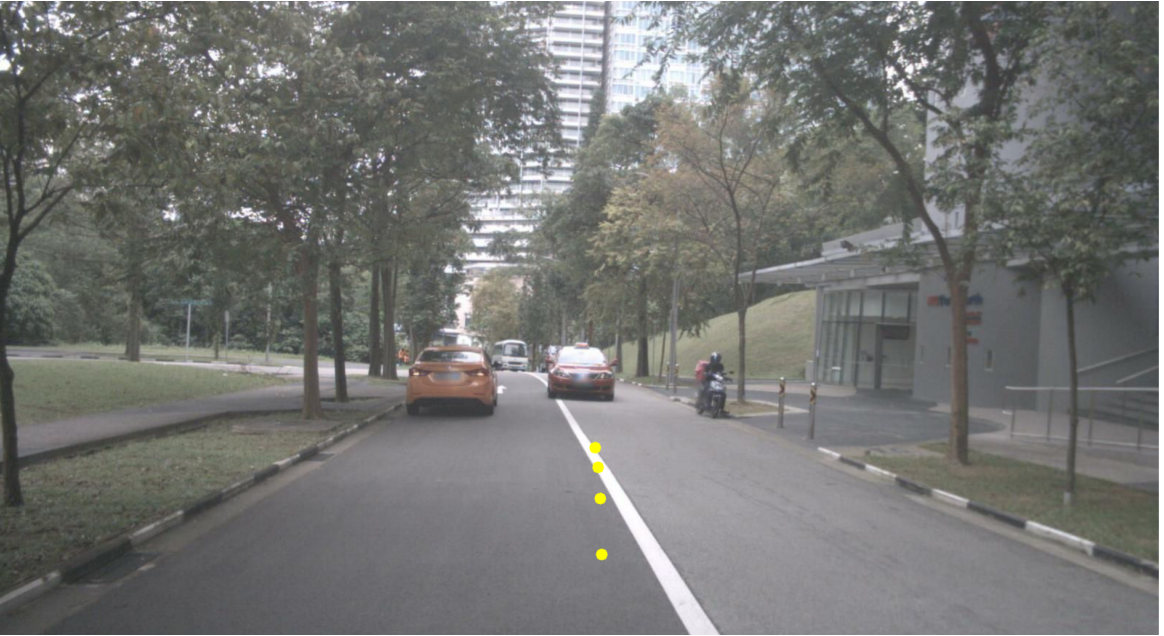}
        \subcaption{Long-tail scenario.}
        \label{fig:case-a}
    \end{subfigure}
    \hfill
    \begin{subfigure}[b]{0.23\textwidth}
        \includegraphics[width=\textwidth]{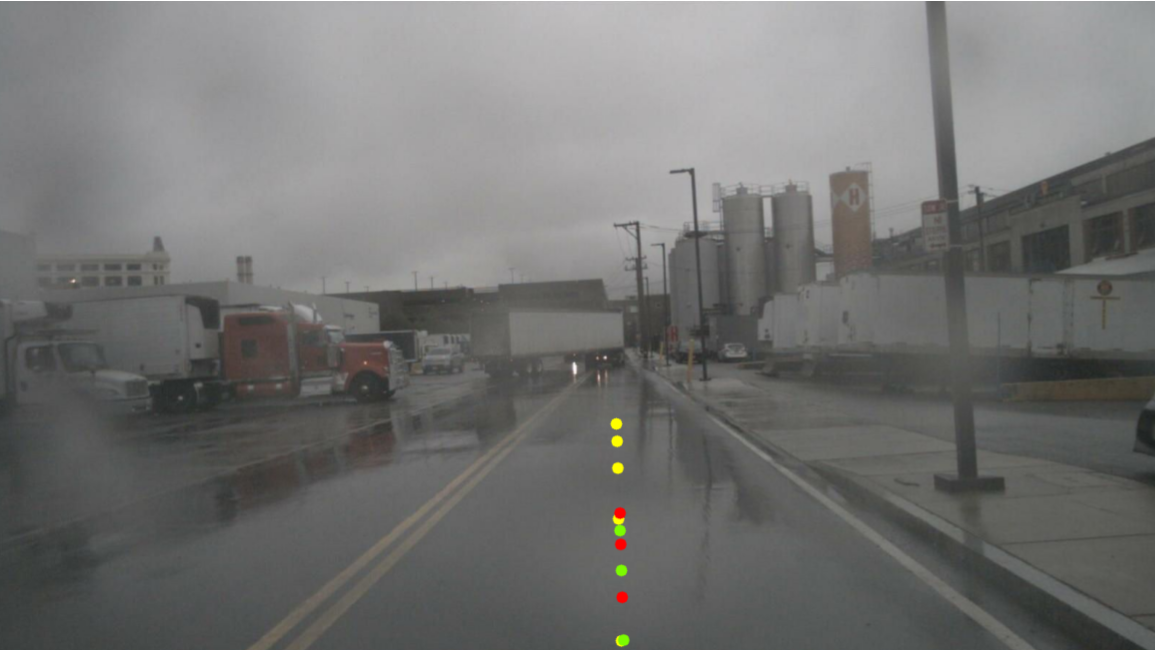} 
        \subcaption{Long-tail scenario.}
        \label{fig:case-d}
    \end{subfigure}
    \hfill
    \begin{subfigure}[b]{0.23\textwidth}
        \includegraphics[width=\textwidth]{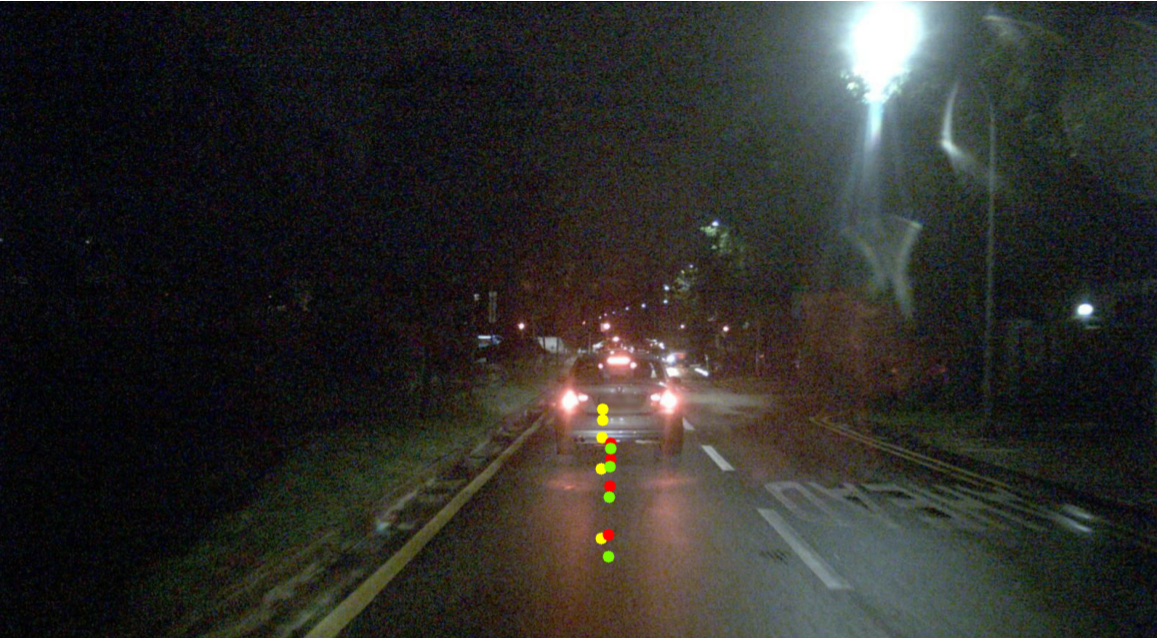}
        \subcaption{H2D scenario.}
        \label{fig:case-b}
    \end{subfigure}
    \hfill
    \begin{subfigure}[b]{0.23\textwidth}
        \includegraphics[width=\textwidth]{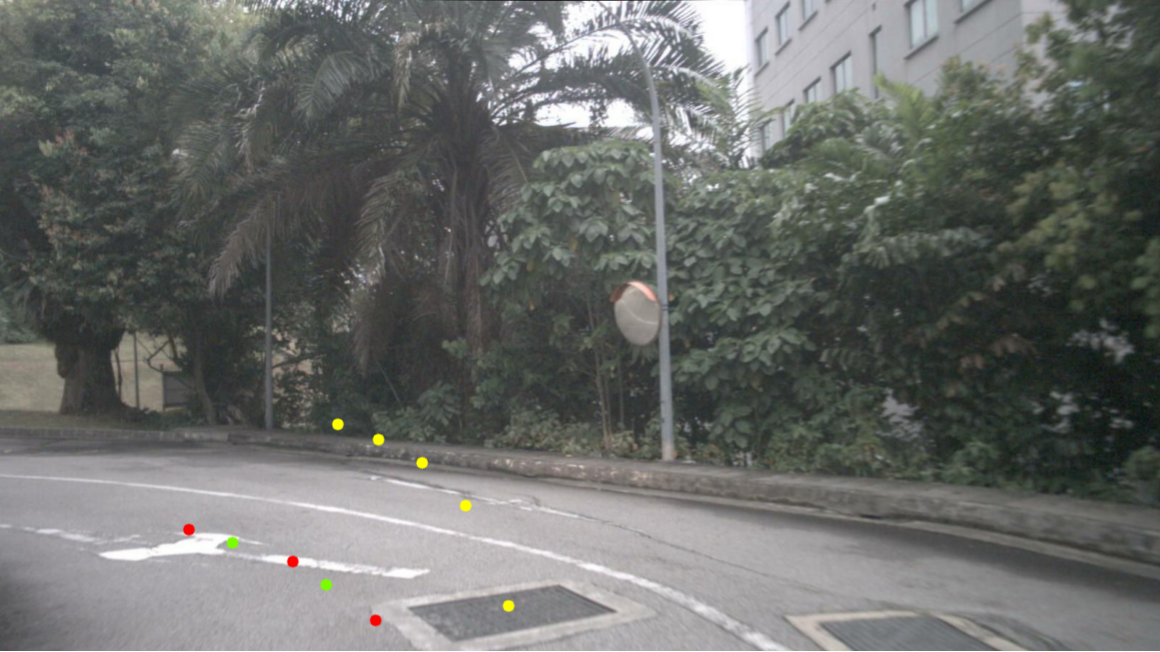}
        \subcaption{H2D scenario.}
        \label{fig:case-c}
    \end{subfigure}
    \vspace{-3mm}
    \caption{Visual comparison of trajectories in challenging scenarios. \textcolor{red}{Ground-truth} trajectories are marked in red, \textcolor{yellow}{baseline} predictions in yellow, and \textcolor{green}{RoboTron-Sim}'s predictions in green.}
    \vspace{-5mm}
    \label{fig:scenario-cases}
\end{figure}

We visualize predicted trajectories from both the baseline and \name{}under ego-pose-free conditions, together with ground truth. In Fig.~\ref{fig:case-a}, when facing an oncoming vehicle crossing lanes and a stationary ego-lane obstacle, \textcolor{green}{RoboTron-Sim} accurately executes a safe stop like the \textcolor{red}{ground truth}, whereas the \textcolor{yellow}{baseline} dangerously continues forward. This demonstrates our model’s superior risk awareness in complex situations. \name{}also yields safer sharp turns (Fig.~\ref{fig:case-c}). These visualizations (and Fig.~\ref{fig:env_comp}) confirm RoboTron-Sim’s effectiveness in safety-critical scenarios.

% , particularly in handling sudden curvature changes.

% In Fig.~\ref{fig:case-d}, the green trajectories exhibit closer alignment with ground truth while maintaining safer lateral deviations compared to the baseline. 
% The night scenario (Fig.~\ref{fig:case-b}) reveals the enhanced perception robustness of RoboTron-Sim, with predicted trajectories that adhere better to road geometry under low-light conditions. 

\vspace{-2mm}
\section{Conclusion}
\label{sec:conclusion}
\vspace{-1mm}
% In this paper, we present \name{}, an approach that enhances real-world driving robustness by strategically leveraging the long-tail and hard-to-driving cases. 
% To our knowledge, we provide the first solution that addresses sim-to-real gaps in MLLMs for autonomous driving. We introduce the HASS dataset and the \name{} model with SPE and I2E, enhancing real-world driving performance in rare high-risk driving scenarios.

% In this paper, we present \name{}, an approach that enhances real-world driving robustness by strategically leveraging long-tail and hard-to-drive cases. To our knowledge, this is the first solution addressing sim-to-real gaps in MLLMs for autonomous driving. We introduce the HASS dataset and the \name{}model with SPE and I2E, which boost real-world performance in rare high-risk driving scenarios.

In this paper, we present RoboTron-Sim, an approach that improves real-world driving robustness by leveraging long-tail and hard-to-drive cases. To our knowledge, this is the first solution addressing sim-to-real gaps in MLLMs for autonomous driving. We introduce the HASS dataset and the \name{}model with SPE and I2E, boosting real-world performance in rare high-risk driving scenarios.

% designs to bridge simulated-real data gaps, enhancing data-scarce scenarios with accessible simulated data.
% \input{ICCV2025-Author-Kit-Feb/sec/appendix}

{
    \small
    \bibliographystyle{ieeenat_fullname}
    % \bibliography{main}

\begin{thebibliography}{10}
\providecommand{\url}[1]{#1}
\csname url@samestyle\endcsname
\providecommand{\newblock}{\relax}
\providecommand{\bibinfo}[2]{#2}
\providecommand{\BIBentrySTDinterwordspacing}{\spaceskip=0pt\relax}
\providecommand{\BIBentryALTinterwordstretchfactor}{4}
\providecommand{\BIBentryALTinterwordspacing}{\spaceskip=\fontdimen2\font plus
\BIBentryALTinterwordstretchfactor\fontdimen3\font minus \fontdimen4\font\relax}
\providecommand{\BIBforeignlanguage}[2]{{%
\expandafter\ifx\csname l@#1\endcsname\relax
\typeout{** WARNING: IEEEtran.bst: No hyphenation pattern has been}%
\typeout{** loaded for the language `#1'. Using the pattern for}%
\typeout{** the default language instead.}%
\else
\language=\csname l@#1\endcsname
\fi
#2}}
\providecommand{\BIBdecl}{\relax}
\BIBdecl

\bibitem{hu2023planning}
Y.~Hu, J.~Yang, L.~Chen, K.~Li, C.~Sima, X.~Zhu, S.~Chai, S.~Du, T.~Lin, W.~Wang \emph{et~al.}, ``Planning-oriented autonomous driving,'' in \emph{Proceedings of the IEEE/CVF conference on computer vision and pattern recognition}, 2023, pp. 17\,853--17\,862.

\bibitem{zhai2023rethinking}
J.-T. Zhai, Z.~Feng, J.~Du, Y.~Mao, J.-J. Liu, Z.~Tan, Y.~Zhang, X.~Ye, and J.~Wang, ``Rethinking the open-loop evaluation of end-to-end autonomous driving in nuscenes,'' \emph{arXiv preprint arXiv:2305.10430}, 2023.

\bibitem{li2024ego}
Z.~Li, Z.~Yu, S.~Lan, J.~Li, J.~Kautz, T.~Lu, and J.~M. Alvarez, ``Is ego status all you need for open-loop end-to-end autonomous driving?'' in \emph{Proceedings of the IEEE/CVF Conference on Computer Vision and Pattern Recognition}, 2024, pp. 14\,864--14\,873.

\bibitem{hwang2024emma}
J.-J. Hwang, R.~Xu, H.~Lin, W.-C. Hung, J.~Ji, K.~Choi, D.~Huang, T.~He, P.~Covington, B.~Sapp \emph{et~al.}, ``Emma: End-to-end multimodal model for autonomous driving,'' \emph{arXiv preprint arXiv:2410.23262}, 2024.

\bibitem{liu2023visual}
H.~Liu, C.~Li, Q.~Wu, and Y.~J. Lee, ``Visual instruction tuning,'' \emph{Advances in neural information processing systems}, vol.~36, pp. 34\,892--34\,916, 2023.

\bibitem{chiang2023vicuna}
W.-L. Chiang, Z.~Li, Z.~Lin, Y.~Sheng, Z.~Wu, H.~Zhang, L.~Zheng, S.~Zhuang, Y.~Zhuang, J.~E. Gonzalez \emph{et~al.}, ``Vicuna: An open-source chatbot impressing gpt-4 with 90\%* chatgpt quality, march 2023,'' \emph{URL https://lmsys. org/blog/2023-03-30-vicuna}, vol.~3, no.~5, 2023.

\bibitem{yu2024merlin}
E.~Yu, L.~Zhao, Y.~Wei, J.~Yang, D.~Wu, L.~Kong, H.~Wei, T.~Wang, Z.~Ge, X.~Zhang \emph{et~al.}, ``Merlin: Empowering multimodal llms with foresight minds,'' in \emph{European Conference on Computer Vision}.\hskip 1em plus 0.5em minus 0.4em\relax Springer, 2024, pp. 425--443.

\bibitem{li2024think2drive}
Q.~Li \emph{et~al.}, ``Think2drive: Efficient reinforcement learning by thinking with latent world model for autonomous driving (in carla-v2),'' in \emph{European Conference on Computer Vision}.\hskip 1em plus 0.5em minus 0.4em\relax Cham: Springer Nature Switzerland, 2024.

\bibitem{caesar2020nuscenes}
H.~Caesar, V.~Bankiti, A.~H. Lang, S.~Vora, V.~E. Liong, Q.~Xu, A.~Krishnan, Y.~Pan, G.~Baldan, and O.~Beijbom, ``nuscenes: A multimodal dataset for autonomous driving,'' in \emph{IEEE Conf. Comput. Vis. Pattern Recog.}, 2020, pp. 11\,621--11\,631.

\bibitem{dosovitskiy2017carla}
A.~Dosovitskiy, G.~Ros, F.~Codevilla, A.~Lopez, and V.~Koltun, ``Carla: An open urban driving simulator,'' in \emph{Conference on robot learning}.\hskip 1em plus 0.5em minus 0.4em\relax PMLR, 2017, pp. 1--16.

\bibitem{shah2018airsim}
S.~Shah, D.~Dey, C.~Lovett, and A.~Kapoor, ``Airsim: High-fidelity visual and physical simulation for autonomous vehicles,'' in \emph{Field and Service Robotics: Results of the 11th International Conference}.\hskip 1em plus 0.5em minus 0.4em\relax Springer, 2018, pp. 621--635.

\bibitem{huang2024making}
Z.~Huang, T.~Tang, S.~Chen, S.~Lin, Z.~Jie, L.~Ma, G.~Wang, and X.~Liang, ``Making large language models better planners with reasoning-decision alignment,'' in \emph{European Conference on Computer Vision}.\hskip 1em plus 0.5em minus 0.4em\relax Springer, 2024, pp. 73--90.

\bibitem{jiang2024senna}
B.~Jiang, S.~Chen, B.~Liao, X.~Zhang, W.~Yin, Q.~Zhang, C.~Huang, W.~Liu, and X.~Wang, ``Senna: Bridging large vision-language models and end-to-end autonomous driving,'' \emph{arXiv preprint arXiv:2410.22313}, 2024.

\bibitem{li2024llava}
B.~Li, Y.~Zhang, D.~Guo, R.~Zhang, F.~Li, H.~Zhang, K.~Zhang, P.~Zhang, Y.~Li, Z.~Liu \emph{et~al.}, ``Llava-onevision: Easy visual task transfer,'' \emph{arXiv preprint arXiv:2408.03326}, 2024.

\bibitem{li2024automated}
Y.~Li, W.~Zhang, K.~Chen, Y.~Liu, P.~Li, R.~Gao, L.~Hong, M.~Tian, X.~Zhao, Z.~Li \emph{et~al.}, ``Automated evaluation of large vision-language models on self-driving corner cases,'' \emph{arXiv preprint arXiv:2404.10595}, 2024.

\bibitem{marcu2024lingoqa}
A.-M. Marcu, L.~Chen, J.~H{\"u}nermann, A.~Karnsund, B.~Hanotte, P.~Chidananda, S.~Nair, V.~Badrinarayanan, A.~Kendall, J.~Shotton \emph{et~al.}, ``Lingoqa: Video question answering for autonomous driving,'' in \emph{Eur. Conf. Comput. Vis.}, 2024.

\bibitem{ding2024holistic}
X.~Ding, J.~Han, H.~Xu, X.~Liang, W.~Zhang, and X.~Li, ``Holistic autonomous driving understanding by bird's-eye-view injected multi-modal large models,'' in \emph{IEEE Conf. Comput. Vis. Pattern Recog.}, 2024, pp. 13\,668--13\,677.

\bibitem{tian2024drivevlm}
X.~Tian, J.~Gu, B.~Li, Y.~Liu, Y.~Wang, Z.~Zhao, K.~Zhan, P.~Jia, X.~Lang, and H.~Zhao, ``Drivevlm: The convergence of autonomous driving and large vision-language models,'' \emph{arXiv preprint arXiv:2402.12289}, 2024.

\bibitem{achiam2023gpt}
J.~Achiam, S.~Adler, S.~Agarwal, L.~Ahmad, I.~Akkaya, F.~L. Aleman, D.~Almeida, J.~Altenschmidt, S.~Altman, S.~Anadkat \emph{et~al.}, ``Gpt-4 technical report,'' \emph{arXiv preprint arXiv:2303.08774}, 2023.

\bibitem{wen2023dilu}
L.~Wen, D.~Fu, X.~Li, X.~Cai, T.~Ma, P.~Cai, M.~Dou, B.~Shi, L.~He, and Y.~Qiao, ``Dilu: A knowledge-driven approach to autonomous driving with large language models,'' \emph{arXiv preprint arXiv:2309.16292}, 2023.

\bibitem{xu2024drivegpt4}
Z.~Xu, Y.~Zhang, E.~Xie, Z.~Zhao, Y.~Guo, K.-Y.~K. Wong, Z.~Li, and H.~Zhao, ``Drivegpt4: Interpretable end-to-end autonomous driving via large language model,'' \emph{IEEE Robotics and Automation Letters}, 2024.

\bibitem{huang2025making}
Z.~Huang, T.~Tang, S.~Chen, S.~Lin, Z.~Jie, L.~Ma, G.~Wang, and X.~Liang, ``Making large language models better planners with reasoning-decision alignment,'' in \emph{European Conference on Computer Vision}.\hskip 1em plus 0.5em minus 0.4em\relax Springer, 2025, pp. 73--90.

\bibitem{cui2024drive}
C.~Cui, Y.~Ma, X.~Cao, W.~Ye, and Z.~Wang, ``Drive as you speak: Enabling human-like interaction with large language models in autonomous vehicles,'' in \emph{Proceedings of the IEEE/CVF Winter Conference on Applications of Computer Vision}, 2024, pp. 902--909.

\bibitem{shao2024lmdrive}
H.~Shao, Y.~Hu, L.~Wang, G.~Song, S.~L. Waslander, Y.~Liu, and H.~Li, ``Lmdrive: Closed-loop end-to-end driving with large language models,'' in \emph{Proceedings of the IEEE/CVF Conference on Computer Vision and Pattern Recognition}, 2024, pp. 15\,120--15\,130.

\bibitem{wang2024omnidrive}
S.~Wang, Z.~Yu, X.~Jiang, S.~Lan, M.~Shi, N.~Chang, J.~Kautz, Y.~Li, and J.~M. Alvarez, ``Omnidrive: A holistic llm-agent framework for autonomous driving with 3d perception, reasoning and planning,'' \emph{arXiv preprint arXiv:2405.01533}, 2024.

\bibitem{liu2024visual}
H.~Liu, C.~Li, Q.~Wu, and Y.~J. Lee, ``Visual instruction tuning,'' \emph{Advances in neural information processing systems}, vol.~36, 2024.

\bibitem{yang2023survey}
Z.~Yang, X.~Jia, H.~Li, and J.~Yan, ``A survey of large language models for autonomous driving,'' \emph{arXiv preprint arXiv:2311.01043}, 2023.

\bibitem{chen2024survey}
L.~Chen, P.~Wu, K.~Chitta, B.~Jaeger, A.~Geiger, and H.~Li, ``End-to-end autonomous driving: Challenges and frontiers,'' \emph{IEEE Transactions on Pattern Analysis and Machine Intelligence}, 2024.

\bibitem{zhang2024survey}
Z.~Zhang, X.~Bo, C.~Ma, R.~Li, X.~Chen, Q.~Dai, J.~Zhu, Z.~Dong, and J.-R. Wen, ``A survey on the memory mechanism of large language model based agents,'' \emph{arXiv preprint arXiv:2404.13501}, 2024.

\bibitem{huang2021bevdet}
J.~Huang, G.~Huang, Z.~Zhu, Y.~Ye, and D.~Du, ``Bevdet: High-performance multi-camera 3d object detection in bird-eye-view,'' \emph{arXiv preprint arXiv:2112.11790}, 2021.

\bibitem{liang2022bevfusion}
T.~Liang, H.~Xie, K.~Yu, Z.~Xia, Z.~Lin, Y.~Wang, T.~Tang, B.~Wang, and Z.~Tang, ``Bevfusion: A simple and robust lidar-camera fusion framework,'' \emph{Advances in Neural Information Processing Systems}, vol.~35, pp. 10\,421--10\,434, 2022.

\bibitem{liu2023bevfusion}
Z.~Liu, H.~Tang, A.~Amini, X.~Yang, H.~Mao, D.~L. Rus, and S.~Han, ``Bevfusion: Multi-task multi-sensor fusion with unified bird's-eye view representation,'' in \emph{2023 IEEE international conference on robotics and automation (ICRA)}.\hskip 1em plus 0.5em minus 0.4em\relax IEEE, 2023, pp. 2774--2781.

\bibitem{gu2023vip3d}
J.~Gu, C.~Hu, T.~Zhang, X.~Chen, Y.~Wang, Y.~Wang, and H.~Zhao, ``Vip3d: End-to-end visual trajectory prediction via 3d agent queries,'' in \emph{Proceedings of the IEEE/CVF Conference on Computer Vision and Pattern Recognition}, 2023, pp. 5496--5506.

\bibitem{gao2020vectornet}
J.~Gao, C.~Sun, H.~Zhao, Y.~Shen, D.~Anguelov, C.~Li, and C.~Schmid, ``Vectornet: Encoding hd maps and agent dynamics from vectorized representation,'' in \emph{Proceedings of the IEEE/CVF conference on computer vision and pattern recognition}, 2020, pp. 11\,525--11\,533.

\bibitem{da2022path}
F.~Da and Y.~Zhang, ``Path-aware graph attention for hd maps in motion prediction,'' in \emph{2022 International conference on robotics and automation (ICRA)}.\hskip 1em plus 0.5em minus 0.4em\relax IEEE, 2022, pp. 6430--6436.

\bibitem{scheel2022urban}
O.~Scheel, L.~Bergamini, M.~Wolczyk, B.~Osi{\'n}ski, and P.~Ondruska, ``Urban driver: Learning to drive from real-world demonstrations using policy gradients,'' in \emph{Conference on Robot Learning}.\hskip 1em plus 0.5em minus 0.4em\relax PMLR, 2022, pp. 718--728.

\bibitem{sadat2020perceive}
A.~Sadat, S.~Casas, M.~Ren, X.~Wu, P.~Dhawan, and R.~Urtasun, ``Perceive, predict, and plan: Safe motion planning through interpretable semantic representations,'' in \emph{Computer Vision--ECCV 2020: 16th European Conference, Glasgow, UK, August 23--28, 2020, Proceedings, Part XXIII 16}.\hskip 1em plus 0.5em minus 0.4em\relax Springer, 2020, pp. 414--430.

\bibitem{gao2022cola}
L.~Gao, Z.~Gu, C.~Qiu, L.~Lei, S.~E. Li, S.~Zheng, W.~Jing, and J.~Chen, ``Cola-hrl: Continuous-lattice hierarchical reinforcement learning for autonomous driving,'' in \emph{2022 IEEE/RSJ International Conference on Intelligent Robots and Systems (IROS)}.\hskip 1em plus 0.5em minus 0.4em\relax IEEE, 2022, pp. 13\,143--13\,150.

\bibitem{yang2023llm4drive}
Z.~Yang, X.~Jia, H.~Li, and J.~Yan, ``Llm4drive: A survey of large language models for autonomous driving,'' \emph{arXiv preprint arXiv:2311.01043}, 2023.

\bibitem{brown2020language}
T.~Brown, B.~Mann, N.~Ryder, M.~Subbiah, J.~D. Kaplan, P.~Dhariwal, A.~Neelakantan, P.~Shyam, G.~Sastry, A.~Askell \emph{et~al.}, ``Language models are few-shot learners,'' \emph{Advances in neural information processing systems}, vol.~33, pp. 1877--1901, 2020.

\bibitem{wu2023policy}
P.~Wu, L.~Chen, H.~Li, X.~Jia, J.~Yan, and Y.~Qiao, ``Policy pre-training for autonomous driving via self-supervised geometric modeling,'' \emph{arXiv preprint arXiv:2301.01006}, 2023.

\bibitem{jia2024bench2drive}
X.~Jia, Z.~Yang, Q.~Li, Z.~Zhang, and J.~Yan, ``Bench2drive: Towards multi-ability benchmarking of closed-loop end-to-end autonomous driving,'' \emph{arXiv preprint arXiv:2406.03877}, 2024.

\bibitem{yang2024oasis}
Z.~Yang, Z.~Zhang, Z.~Zheng, Y.~Jiang, Z.~Gan, Z.~Wang, Z.~Ling, J.~Chen, M.~Ma, B.~Dong, \emph{et~al.}, ``Oasis: Open agent social interaction simulations with one million agents,'' \emph{arXiv preprint arXiv:2411.11581}, 2024.

\bibitem{sun2020waymo}
P.~Sun, H.~Kretzschmar, X.~Dotiwalla, A.~Chouard, V.~Patnaik, P.~Tsui, J.~Guo, Y.~Zhou, Y.~Chai, B.~Caine \emph{et~al.}, ``Scalability in perception for autonomous driving: Waymo open dataset,'' in \emph{Proceedings of the IEEE/CVF conference on computer vision and pattern recognition}, 2020, pp. 2446--2454.

\bibitem{jain2021autonomy}
A.~Jain, L.~Del~Pero, H.~Grimmett, and P.~Ondruska, ``Autonomy 2.0: Why is self-driving always 5 years away?'' \emph{arXiv preprint arXiv:2107.08142}, 2021.

\bibitem{huang2024drivemm}
Z.~Huang, C.~Feng, F.~Yan, B.~Xiao, Z.~Jie, Y.~Zhong, X.~Liang, and L.~Ma, ``Drivemm: All-in-one large multimodal model for autonomous driving,'' \emph{arXiv preprint arXiv:2412.07689}, 2024.

\bibitem{yan2024robomm}
F.~Yan, F.~Liu, L.~Zheng, Y.~Zhong, Y.~Huang, Z.~Guan, C.~Feng, and L.~Ma, ``Robomm: All-in-one multimodal large model for robotic manipulation,'' \emph{arXiv preprint arXiv:2412.07215}, 2024.

\bibitem{zhong2025p3nav}
Y.~Zhong, C.~Feng, F.~Yan, F.~Liu, L.~Zheng, and L.~Ma, ``P3nav: A unified framework for embodied navigation integrating perception, planning, and prediction,'' \emph{arXiv preprint arXiv:2503.18525}, 2025.

\bibitem{jiang2023vad}
B.~Jiang, S.~Chen, Q.~Xu, B.~Liao, J.~Chen, H.~Zhou, Q.~Zhang, W.~Liu, C.~Huang, and X.~Wang, ``Vad: Vectorized scene representation for efficient autonomous driving,'' in \emph{Proceedings of the IEEE/CVF International Conference on Computer Vision}, 2023, pp. 8340--8350.

\end{thebibliography}

}

\clearpage 
\appendix  % 开始附录部分
% \section{Appendix}

\section{Experimental Details}
\subsection{Datasets}
RoboTron-Sim is trained using a hybrid data strategy combining:
\begin{itemize}
    \item \textbf{Real-world Data}: 28,130 samples from nuScenes\cite{caesar2020nuscenes}.
    
    \item \textbf{Simulated Data}: 47,553 purpose-built samples from our Hard-case Augmented Synthetic Scenarios(HASS) dataset, generated in CARLA simulator\cite{dosovitskiy2017carla}, designed to address the inherent imbalance in real-world data distribution. While the dataset covers a broad range of driving situations, it places particular emphasis on addressing challenging cases, including H2D scenarios and Long-Tail scenarios. Partial results are illustrated in Figure~\ref{fig:HASS}.
    
% 
    
    % 
    % \item \textbf{Test Set}: 6,019 samples from nuScenes validation split, ensuring evaluation consistency with real-world benchmarks
\end{itemize}

\subsection{Evaluation Metrics}

Following BEV-Planner~\cite{li2024ego}, we evaluate via L2 Distance, Collision Rate, and Boundary Violation Rate. 

\begin{itemize}
    \item \textbf{Trajectory Accuracy (L2 Distance)}:  
    \begin{equation}
        L2 = \frac{1}{T}\sum_{t=1}^{T} \| \hat{p}_t - p^{gt}_t \|_2
    \end{equation}
    where $\hat{p}_t$ and $p^{gt}_t$ denote the predicted and ground-truth positions at timestep $t$ over a $T=3s$ horizon.  

    \item \textbf{Safety Metrics (Collision Rate)}:  
        
        Computes the percentage of predicted trajectories that result in collisions with other agents or obstacles.  
        \begin{equation}
    Collision = \frac{1}{T} \sum_{t=1}^{T} \frac{N_{\text{collision}, t}}{N_{\text{total}, t}} \times 100\%
        \end{equation}

        where $N_{\text{collision}}$ is the number of predicted trajectories leading to collisions, and $N_{\text{total}}$ is the total number of evaluated trajectories at timestep $t$ over a $T=3s$ horizon.  

\end{itemize}

\begin{itemize}
    \item \textbf{Boundary Violation Rate}:  
    \begin{equation}
        Boundary = \frac{1}{T} \sum_{t=1}^{T} \frac{N_{\text{violation}, t}}{N_{\text{total}, t}} \times 100\%
    \end{equation}
    where $N_{\text{violation}}$ counts trajectories exceeding road boundaries, and $N_{\text{total}}$ is the total evaluated trajectories at timestep $t$ over $T=3s$. Calculated by comparing ego segmentation masks with drivable area labels.
    
\end{itemize}

\section{More Results}
\subsection{Robustness of HASS}
We investigate the performance trend divergence between simulated data augmentation and real data-only scenarios across multiple orders of magnitude in real data volume in RoboTron-Sim, with quantitative comparisons presented in Table~\ref{tab:traing_data_ablation_sup} and Table~\ref{tab:traing_data_ablation2_supp}. Table~\ref{tab:traing_data_ablation_sup} presents quantitative results with full simulated data integration, while Table~\ref{tab:traing_data_ablation2_supp} provides detailed metrics when trained without any simulated data, using real-world data exclusively. The experimental results demonstrate enhanced stability of overall performance through simulated data augmentation.

% 少样本实验的表格
\begin{table}[t]
% \small
\centering
\addtolength{\tabcolsep}{0pt}
\begin{tabular}{cc|cc} %{l|ccc|c|ccc|c}
\toprule

nuScenes & HASS & L2(m) & Collision(\%)  \\
 
\midrule
0\% & 100\% & 1.24 & 0.99 \\
10\% & 100\% & 0.87 & 0.89  \\
25\% & 100\% & 0.67 & 0.64 \\
50\% & 100\% & 0.63 & 0.57 \\
75\% & 100\% & 0.61 & 0.54 \\
100\% & 100\% & 0.56 & 0.58 \\

\bottomrule
\end{tabular}
% \vspace{4pt}
\caption{Performance variation with nuScenes blending ratio under full HASS integration.
% We compare the model's performance when trained on a single dataset and on all datasets combined.
}
\label{tab:traing_data_ablation_sup}
\end{table}

\begin{table}[t]
% \small
\centering
\addtolength{\tabcolsep}{0pt}
\begin{tabular}{cc|cc} %{l|ccc|c|ccc|c}
\toprule

nuScenes & HASS & L2(m) & Collision(\%)  \\
 
\midrule
10\% & 0\% & 2.41 & 3.22   \\
25\% & 0\% & 1.24 & 1.46  \\
50\% & 0\% & 1.15 & 1.54  \\
75\% & 0\% & 1.03 & 1.03  \\
100\% & 0\% & 0.72 & 0.67 \\

\bottomrule
\end{tabular}
% \vspace{4pt}
\caption{
Performance scaling with nuScenes blending ratio (HASS Excluded).
}
\label{tab:traing_data_ablation2_supp}
\end{table}

\subsection{Effectiveness of HASS}
We generate two distinct datasets based on nuScenes scenarios: General Augmented Synthetic Scenarios (GASS) for common driving conditions and Hard-case Augmented Synthetic Scenarios (HASS) for challenging situations, aiming to investigate which synthetic data generation mechanisms yield more meaningful performance improvements. The evaluation results categorized by individual scenarios are presented in Table~\ref{tab:sim_scenario_comparison_supp}, while the aggregated metrics for H2D scenarios (Night+Turn+Rain) are summarized in Table~\ref{tab:sim_scenario_comparison3_5}.

\begin{table*}[t]
    % \small % 这里选择小一号字体，不是最小
    \centering
    \begin{tabularx}{\textwidth}{l|X X|X X|X X|X X|X X|X X}
        \toprule
        \multirow{2}{*}{\textbf{Data}} 
        & \multicolumn{2}{c|}{\textbf{Day}} 
        & \multicolumn{2}{c|}{\textbf{Night}} 
        & \multicolumn{2}{c|}{\textbf{Straight}} 
        & \multicolumn{2}{c|}{\textbf{Turn}} 
        & \multicolumn{2}{c|}{\textbf{Sunny}} 
        & \multicolumn{2}{c}{\textbf{Rainy}} \\
        \cmidrule(lr){2-3} \cmidrule(lr){4-5} \cmidrule(lr){6-7}
        \cmidrule(lr){8-9} \cmidrule(lr){10-11} \cmidrule(l){12-13}
        & L2 & Col 
        & L2 & Col 
        & L2 & Col
        & L2 & Col
        & L2 & Col
        & L2 & Col \\
        \midrule
        nuScenes & 0.59 & 0.50 & 1.40 & 2.71 & 0.59 & 0.55 & 1.32 & 1.80 & 0.64 & 0.63 & 1.15 & 0.81 \\
        nuScenes + GASS & 0.55 & 0.42 & 1.00 & 2.53 & 0.50 & 0.49 & 1.21 & 1.89 & 0.52 & 0.58 & 0.99 & 0.79 \\
        nuScenes + HASS & 0.54 & 0.47 & 0.81 & 1.56 & 0.55 & 0.52 & 0.64 & 1.01 & 0.56 & 0.64 & 0.56 & 0.32 \\
        \bottomrule
    \end{tabularx}
    \caption{Performance comparison across various training data in each scenario.}
    \label{tab:sim_scenario_comparison_supp}
\end{table*}

\begin{table}[t]
\centering
\scalebox{0.90}{% 缩放因子设为 0.85
\begin{minipage}{\textwidth} % 保持原始文本宽度
\small
\setlength{\tabcolsep}{3pt}
\begin{tabular}{l|c c|c c}
\toprule
\multirow{2}{*}{Training Data} & 
\multicolumn{2}{c|}{E2D} & 
\multicolumn{2}{c}{H2D} \\
\cmidrule(lr){2-3} \cmidrule(l){4-5}
& L2 (m) & Collision (\%) & L2 (m) & Collision (\%) \\
\midrule
nuScenes        & 0.61 & 0.56 & 1.29 & 1.77 \\
nuScenes + GASS & 0.52 & 0.50 & 1.07 & 1.74 \\
nuScenes + HASS & 0.55 & 0.54 & 0.67 & 0.96 \\
\bottomrule
\end{tabular}
\end{minipage}
}
\vspace*{-4pt} 
\caption{Performance comparison in H2D and E2D scenarios.}
\label{tab:sim_scenario_comparison3_5}
\end{table}

\subsection{Model Generalization}
To verify the model generalization in the \textbf{planning task}, we further evaluate model performance on the NAVSIM (NV) benchmark using the predictive driver model score (PDMS), which is based on five factors: no at-fault collisions (NC), drivable area compliance (DAC), time-to-collision (TTC), comfort (Comf.), and ego progress (EP). Table~\ref{tab:generalization} demonstrates that RoboTron-Sim delivers comparable or superior performance compared to existing methods, with the integration of HASS achieving a PDMS of 85.6 and setting a new SOTA result on NV benchmark.

\begin{table}[t]
% \centering
\small
\renewcommand{\arraystretch}{1.0} % 减小行间距（默认值是 1.0）
\setlength{\tabcolsep}{1.5pt} % 减少列间距（默认值可能是 6pt）
\scalebox{0.95}{ % 不做全局缩放，让线宽维持在 \textwidth 范围
\begin{tabularx}{0.5\textwidth}{@{}lccccccc@{}} % 使用 '@{}' 去除两端多余空白
\toprule
Method               & Data       & NC↑   & DAC↑   & TTC↑   & Comf.↑ & EP↑   & PDMS↑ \\
\midrule
Human               & -          & 100.0 & 100.0  & 100.0  & 99.9   & 87.5  & 94.8  \\
\midrule
Ego-MLP             & NV         & 93.0  & 77.3   & 83.6   & \textbf{100.0}  & 62.8  & 65.6  \\
UniAD               & NV         & 97.8  & 91.9   & 92.9   & \textbf{100.0}  & 78.8  & 83.4  \\
ParaDrive           & NV         & 97.9  & 92.4   & 93.0   & 99.8   & 79.3  & 84.0  \\
RoboTron-Sim            & NV         & 98.0  & 93.0   & 93.3   & 99.8   & 79.9   & 84.6  \\
RoboTron-Sim            & NV+HASS    & \textbf{98.2} & \textbf{93.6} & \textbf{93.8} & 99.9 & \textbf{81.1} & \textbf{85.6} \\
\bottomrule
\end{tabularx}}
\caption{Performance on NAVSIM benchmark, \textdagger indicates that RoboTron-Sim is trained without HASS.}
\label{tab:generalization}
\end{table}

% \vspace{-3.1ex} % 减少表格与文字间距
We also explore the robustness of the model on the \textbf{VQA task}. The VQA data curated for HASS encompasses three categories of questions: (1) Descriptive questions, such as ``What is the color of the traffic light ahead?", are answered directly using data generated by the simulator; (2) Hypothetical questions, such as ``If you turn right at this intersection, what would you encounter?", are annotated using GPT-4o based on environment visuals and predefined rules; (3) Reasoning questions, such as ``Why are you slowing down here?", are generated by GPT-4o based on driving videos and trajectories to enhance the understanding of vehicle behavior. We conduct separate validations on the BDD-X and LingoQA datasets. As shown in Table~\ref{tab:generalization_VQA}, with HASS integration, RoboTron-Sim achieves SOTA performance on both benchmarks (e.g., improving METEOR from 52.23 to 56.30 on BDD-X, and increasing CIDEr from 61.3 to 62.2 on LingoQA).

% achieves new SOTA on both datasets.

% consistently outperforming other models.

% \vspace{-2ex} % 减少表格与文字间距
\begin{table}[t]
\centering
% \small
% \caption{模型在不同数据集上的性能对比}
\label{tab:model_performance}
\resizebox{0.48\textwidth}{!}{ % 控制为半页宽度
\begin{tabular}{@{}lcccc@{}}
\toprule
Method & Data & BLEU & METEOR & CIDEr \\
\midrule
% \multicolumn{5}{l}{\textbf{BDD-X 数据集}} \\
QwenVL & BDD-X & 25.89 & 46.54 & 19.91 \\
LLaVA-1.5 & BDD-X & 25.97 & 45.08 & 21.62 \\
Senna & BDD-X & 31.04 & 50.44 & 34.31 \\
RoboTron-Sim & BDD-X & 32.54 & 52.23 & 37.19 \\
RoboTron-Sim & BDD-X+HASS & \textbf{33.25} & \textbf{56.30} & \textbf{38.17} \\  % 修正合并标识

\midrule
% \multicolumn{5}{l}{\textbf{LingoQA 数据集}} \\
LLaVA & LingoQA & 12.5 & 18.5 & 57.0 \\
Vicuna-7B & LingoQA & 10.1 & 15.2 & 51.0 \\
BLIP-2 & LingoQA & 13.0 & 17.4 & 60.1 \\
LingoQA & LingoQA & 15.0 & 18.6 & 59.5 \\
RoboTron-Sim & LingoQA & 15.5 & 18.5 & 61.3 \\
RoboTron-Sim & LingoQA+HASS & \textbf{16.6} & \textbf{19.0} & \textbf{62.2} \\ % 修正合并标识
\bottomrule
\end{tabular}%
}
\caption{Performance on NAVSIM benchmark, \textdagger indicates that RoboTron-Sim is trained without HASS.}
\label{tab:generalization_VQA}
\end{table}

\begin{table*}[t]
\centering
\addtolength{\tabcolsep}{5pt} % 适当减少列间距
\begin{tabular}{l|c|ll}
\toprule
\multirow{2}{*}{Method} & \multirow{2}{*}{Data} & 
\multicolumn{2}{c}{L2 Distance (m)} \\
\cmidrule(lr){3-4}
& & E2D & H2D \\
\midrule

\multirow{2}{*}{VAD} 
 & nuScenes & 0.78 & 0.88 \\
 & nuScenes + HASS & 
 0.76$\color{blue}({\downarrow{2.5\%}})$ & 
 0.87$\color{blue}({\downarrow{1.1\%}})$ \\
\midrule

\multirow{2}{*}{MLLM} 
 & nuScenes & 1.00 & 1.23 \\
 & nuScenes + HASS & 0.91$\color{blue}({\downarrow{9.0\%}})$ & 1.14$\color{blue}({\downarrow{7.3\%}})$ \\
\midrule

\multirow{2}{*}{RoboTron-Sim} 
 & nuScenes & 0.61 & 1.29 \\
 & nuScenes + HASS & 
 0.57$\color{blue}({\downarrow{6.6\%}})$ & 
 0.67$\color{blue}({\downarrow{48.1\%}})$ \\
\bottomrule
\end{tabular}
\vspace*{-4pt} 
\caption{L2 Distance performance gains of HASS across different models in E2D and H2D scenarios. To rigorously evaluate the model's inherent capability to comprehend dynamic environments without relying on ego-pose dependencies, we conducted ablation studies by removing ego-pose inputs from both MLLM and RoboTron-Sim architectures.}
\label{tab:scenario_comparison}
\end{table*}

\begin{table}[t]
\centering
\resizebox{0.48\textwidth}{!}{ % 保持半页宽
\setlength{\tabcolsep}{4.5pt} % 压缩列间距
\begin{tabular}{@{}l|c|ccc|ccc@{}}
\toprule
\multirow{2}{*}{Model} & 
\multirow{2}{*}{Latency} & 
\multicolumn{3}{c|}{E2D} & 
\multicolumn{3}{c}{H2D} \\
\cmidrule(lr){3-5} \cmidrule(l){6-8}
& & Day & Straight & Sunny & Night & Turn & Rainy \\
\midrule
VAD & 115.3ms & 0.77 & 0.78 & 0.78 & 0.94 & 0.87 & 0.83 \\
RoboTron-Sim-7B & 612.8ms & 0.54 & 0.55 & 0.56 & 0.81 & 0.64 & 0.56 \\
RoboTron-Sim-0.5B & 141.4ms & 0.57 & 0.62 & 0.60 & 0.81 & 0.69 & 0.64 \\

\bottomrule

\end{tabular}%
}
% \vspace{-2ex} % 减少表格与文字间距
\caption{Comparison of deployment costs.}
\label{table:deployment_costs}

\end{table}

\subsection{Model Compatibility}

We conduct a comparative analysis of three models: VAD~\cite{jiang2023vad} (representing classical end-to-end models), LLaVA-OneVision~\cite{li2024llava} (as a representative multimodal large language model), and our RoboTron-Sim, evaluating their performance gains when augmenting real-world data with simulated data. To systematically investigate model compatibility with simulated data augmentation, we conduct cross-architecture evaluations on L2 distance. 
As evidenced in Table~\ref{tab:scenario_comparison}, VAD exhibits fundamental compatibility limitations, with marginal L2 reductions (↓2.5\% E2D, ↓1.1\% H2D). Although MLLM demonstrates preliminary compatibility, showing gradual improvements, the gains remain constrained in the hard cases (↓9.0\% E2D, ↓7.3\% H2D). In stark contrast, our RoboTron-Sim achieves breakthrough enhancements (↓48.1\% ) in H2D case while maintaining stable performance in E2D case. This empowers knowledge transfer from synthetic domains while preserving real-world physical constraints, unlocking the model's untapped potential.

\subsection{Deployment Costs}
We compare the key deployment metrics for the models on RTX-4090, as shown in Table~\ref{table:deployment_costs}. It shows that RoboTron-Sim is applicable to smaller models like RoboTron-Sim-0.5B (replacing the LLM from Qwen2-7B to Qwen1.5-0.5B), achieving comparable performance to RoboTron-Sim-7B and exhibiting deployment efficiency akin to traditional end-to-end model. This alignment of low deployment costs and performance improvement makes RoboTron-Sim practical for a wide range of real-world applications.

\begin{figure*}[htbp]
    \centering
    \foreach \i in {1,...,8} { % 根据图片个数自动调整
        \begin{subfigure}[b]{0.48\textwidth}
            \includegraphics[width=\linewidth]{I\i.png} 
            \subcaption{Scenario \i}
            % \label{fig:case-\i}
        \end{subfigure}%
        \ifnum\i=2 \hfill \par \fi % 每行两个图，第二张后换行
        \ifnum\i=4 \hfill \par \fi % 第四张后换行
    }
    \caption{Visualization of HASS. 
    }
    \label{fig:HASS}
\end{figure*}

\section{Visualization}
\subsection{In Hard-to-Drive(H2D) Scenarios}
We conduct trajectory visualization comparisons among the baseline method, RoboTron-Sim, and ground truth (GT) using representative long-tail cases from the nuScenes test set (including turn, night, and similar challenging scenarios), as shown in Figure~\ref{fig:scenario-cases2}.

\subsection{In Long-Tail Scenarios}
We conduct trajectory visualization comparisons among the baseline method, RoboTron-Sim, and ground truth (GT) using representative long-tail cases from the nuScenes test set (including lane invasion, temporary parking ahead, and similar challenging scenarios), as shown in Figure~\ref{fig:scenario-cases}.

\begin{figure*}[htbp]
    \centering
    \foreach \i in {1,...,8} { % 根据图片个数自动调整
        \begin{subfigure}[b]{0.48\textwidth}
            \includegraphics[width=\linewidth]{H2D\i.png} 
            \subcaption{Scenario \i}
            % \label{fig:case-\i}
        \end{subfigure}%
        \ifnum\i=2 \hfill \par \fi % 每行两个图，第二张后换行
        \ifnum\i=4 \hfill \par \fi % 第四张后换行
    }
    \caption{Visual comparison of planning trajectories in H2D scenarios. 
    \textcolor{red}{Ground-truth} trajectories are marked in red, 
    \textcolor{yellow}{baseline} predictions in yellow, and 
    \textcolor{green}{RoboTron-Sim}'s predictions in green.}
    \label{fig:scenario-cases2}
\end{figure*}

\begin{figure*}[htbp]
    \centering
    \foreach \i in {1,...,6} { % 根据图片个数自动调整
        \begin{subfigure}[b]{0.48\textwidth}
            \includegraphics[width=\linewidth]{LT\i.png} 
            \subcaption{Scenario \i}
            % \label{fig:case-\i}
        \end{subfigure}%
        \ifnum\i=2 \hfill \par \fi % 每行两个图，第二张后换行
        \ifnum\i=4 \hfill \par \fi % 第四张后换行
    }
    \caption{Visual comparison of planning trajectories in Long-Tail scenarios. 
    \textcolor{red}{Ground-truth} trajectories are marked in red, 
    \textcolor{yellow}{baseline} predictions in yellow, and 
    \textcolor{green}{RoboTron-Sim}'s predictions in green.}
    \label{fig:scenario-cases}
\end{figure*}
% 传统方法和大模型对比

% 不同场景下的GASS/HASS对比

% \begin{table}[!t]
% \centering
% \scalebox{0.90}{% 缩放因子设为 0.85
% \begin{minipage}{\textwidth} % 保持原始文本宽度
% \small
% \setlength{\tabcolsep}{3pt}
% \begin{tabular}{l|c c|c c}
% \toprule
% \multirow{2}{*}{Training Data} & 
% \multicolumn{2}{c|}{E2D} & 
% \multicolumn{2}{c}{H2D} \\
% \cmidrule(lr){2-3} \cmidrule(l){4-5}
% & L2 (m) & Collision (\%) & L2 (m) & Collision (\%) \\
% \midrule
% nuScenes        & 0.61 & 0.56 & 1.29 & 1.77 \\
% nuScenes + GASS & 0.52 & 0.50 & 1.07 & 1.74 \\
% nuScenes + HASS & 0.55 & 0.54 & 0.67 & 0.96 \\
% \bottomrule
% \end{tabular}
% \end{minipage}
% }
% \vspace*{-4pt} 
% \caption{Performance comparison in H2D and E2D scenarios.}
% \label{tab:sim_scenario_comparison3_5}
% \end{table}

\def\ie{{\em i.e.}}
\def\eg{{\em e.g.}}
\def\etal{{\em et al.}}
\def\etc{{\em etc.}}
\newcommand{\cmtt}[1]

\end{document}